# Human–AI Co-Embodied Intelligence for Scientific Experimentation and Manufacturing

Xinyi Lin[1,3], Yuyang Zhang[1,3], Yuanhang Gan[1,3], Juntao Chen[1], Hao Shen[1], Yichun He[1], Lijun Li[2], Ze Yuan[2], Shuang Wang[2], Chaohao Wang[2], Rui Zhang[2], Na Li[1], Jia Liu[1,*]

[1] John A. Paulson School of Engineering and Applied Sciences, Harvard University, Cambridge, MA, USA.
[2] UltraReality Technology Limited, Mountain View, CA, USA.
[3] These authors contributed equally.
[*] e-mail: jia_liu@seas.harvard.edu.

## Abstract

Scientific experimentation and manufacturing rely on prolonged protocol development and complex, multi-step implementation, which require continuous human expertise for precise execution and decision-making, limiting interpretability and scalability. Here, we introduce human-artificial intelligence (AI) co-embodied intelligence, a new form of physical AI that unites human researchers, agentic AI, and wearable hardware. In this paradigm, humans provide precise execution, while agentic AI contributes contextual reasoning, adaptive planning, and analysis. The wearable hardware facilitates seamless communication between humans and AI by streaming experimentation and manufacturing context to AI and rendering real-time feedback to humans *in situ*. We instantiate this paradigm in a microfabrication cleanroom, leading to the agentic-physical experimentation (APEX) system which understands fabrication procedures with accuracy 51% higher than state-of-the-art multimodal large language models/vision language models (LLMs/VLMs), detects and corrects fabrication errors in real-time, and transfers procedural expertise to novice users. Critically, APEX system enables the co-development of fabrication protocols in cleanrooms, overcoming the incompatibility of elastomeric materials in standard microfabrication processes and enabling previously unattainable fabrication outcomes, as demonstrated by the wafer-scale realization of brain-level soft neural probes capable of single-unit-resolution neural recording. These results establish the human-AI co-embodied intelligence that extends agentic reasoning beyond computation into the physical domain, transforming scientific experimentation and manufacturing into autonomous, traceable, interpretable and scalable processes.

## Main

Scientific experimentation and manufacturing rely on complex, multi-step experimental and fabrication procedures that demand precise coordination among human perception, reasoning, and execution[1-10]. In practice, even when standard operating procedures (SOPs) are explicitly documented, successful execution often depends on tacit knowledge such as timing sensitivity, manual handling nuances, and context-dependent judgment that are difficult to formalize or transfer across users and facilities. Despite rapid advances in machine learning and automation, conventional machine learning models primarily operate as pattern recognizers optimized for fixed datasets and narrowly defined tasks[11,12], and therefore fail to participate in experimental and manufacturing workflows due to limited ability to reason and generalize[12-15]. As a result, they remain confined to virtual domains[16,17], analyzing data[18], predicting outcomes[19-21], or generating code[22,23], rarely capturing the procedural and causal structure that links physical actions to outcomes during execution, while real-world experimentation and manufacturing continue to depend heavily on human execution and supervision. In many experimental settings, critical failures arise from subtle deviations during physical execution that are transient, irreversible, and cannot be recovered through *post hoc* analysis. This gap between digital intelligence and physical execution has become a bottleneck to achieving accessibility, reproducibility, and scalability of scientific experimentation and manufacturing.

Recent advances in multimodal large language models/vision language models (LLMs/VLMs) have introduced elements of reasoning and generalization, enabling artificial intelligence (AI) agents that can develop code, plan experiments, interpret data, and analyze the results[24-30]. These advances provide a foundation for moving beyond purely reactive models toward systems that can reason about procedures and context. These models are widely deployed to automate labor-intensive tasks such as gene-editing design[28], spike-sorting[31], behavior analysis[32], etc. However, most current applications remain computational, focusing on analysis or planning while leaving physical execution largely unaddressed. Integration of these agentic AI systems with robotic systems has led to initial results in autonomous laboratories that plan, execute, and interpret experiments[33-36]. Nevertheless, these embodied systems still cannot be applied to improve the efficiency of complex experimentation and manufacturing, which is largely limited by the incapacity of current robotic systems to perform long-horizon, dexterous, and safety-critical control tasks (e.g., cleanroom fabrication). Thus, the gap between reasoning and physical action persists[33,37-39].

To bridge this gap, we propose the concept of human–AI co-embodied intelligence, which is a new form of physical AI that unites human researchers or operators, agentic AI, and wearable hardware into an integrated system for scientific experimentation and intelligent manufacturing. In this paradigm, humans provide long-horizon execution and delicate motor control that current robotic systems cannot achieve[33]. Agentic AI contributes high-level planning, contextual reasoning, and memory, analyzing ongoing procedures, evaluating outcomes, and generating adaptive guidance. The wearable hardware continuously visualizes both human actions and experimental and manufacturing contexts, allowing the agentic AI to provide continuous analysis of execution, real-time feedback, and corrective instructions. By coupling intelligent reasoning directly with physical execution, human–AI co-embodiment enables humans and AI agents to iteratively co-develop procedural logic as experiments unfold, forming a self-evolving, interpretable, and adaptive system for real-world scientific experimentation, and manufacturing.

As a demonstration, we present APEX system, an agentic-physical experimentation framework that integrates agentic reasoning with real-world perception and physical execution through mixed reality (MR) interaction with humans. Rather than replacing human expertise, APEX system functions as an intelligent collaborator that continuously observes and interprets human actions within complex, multi-step experimental and manufacturing workflows. By maintaining procedural context over extended tasks, APEX system enables real-time co-analysis of execution, shifting experimental understanding from *post hoc* interpretation to *in situ* reasoning. As a result, experimentation and manufacturing are no longer treated as the execution and manual optimization of static protocols, but as adaptive processes in which procedural logic emerges through continuous interaction between human action, agentic reasoning, and physical outcomes.

To validate the framework, we implement APEX system in a microfabrication cleanroom for micro-electro-mechanical systems (MEMS) device development and fabrication, which (1) achieves context-aware procedural reasoning with accuracy exceeding state-of-the-art multimodal LLMs; (2) collaborates with human experts to detect and correct fabrication errors in real-time; (3) transfers procedural expertise to novices, enabling rapid skill acquisition, consistent reproducibility, and scalable manufacturing. Most importantly, we demonstrated that APEX system is capable of co-developing new fabrication protocols to incorporate unconventional materials into the standard device fabrication protocols. As an example, APEX system generates and optimizes the protocol that allows multiple elastomers to be incorporated for the direct fabrication of brain-level soft neural probes for stable single-unit resolution neural recording, which were previously considered impractical to fabricate using conventional approaches or restricted to a narrow subset of microfabrication-compatible elastomeric materials. These results demonstrate the human–AI co-embodied intelligence in which reasoning extends beyond computation into the physical domain, bringing opportunities for a new era of interpretable, scalable, and autonomous scientific experimentation and intelligent manufacturing.

## Results

### APEX system framework

To enable real-time guidance and collaboration with humans during complex fabrication processes in the cleanroom[40,41], APEX system is equipped with the following core capabilities: (1) understanding human operations in the context of fabrication protocol to provide guidance; (2) analyzing procedural information (e.g., step sequences, timing, tool states, and material conditions) based on domain knowledge to detect errors; and (3) reasoning about experiment design, operations, and outcomes to improve existing protocols and develop new protocols.

To enable these functions, APEX system integrates multimodal perception, agentic reasoning, and human–AI collaboration into an end-to-end framework (**Fig. 1a**). Specifically, the APEX system consists of (1) a real-time MR hardware platform that provides real-time multimodal perception for agentic reasoning and immersive MR rendering for human–AI interaction, and (2) a multi-agent reasoning framework composed of multimodal LLMs/VLMs-driven, in-context-learning-based AI agents that enables planning, perception, reasoning, and adaptive feedback (**Fig. 1b**). The core feature of the multi-agent reasoning framework is the evolving capability that enables the agents to evolve both within each experiment and across multiple experiments. Within each experiment, the agents autonomously maintain evolving memories of fabrication operations as the experiment proceeds. The memories retain detail on recent and relevant operations while condensing past and less relevant operations, providing a real-time adaptive context for the agents to understand, guide, and correct human operations. Across multiple experiments, the agents summarize the protocol design, operation log and experiment experiences and add to their memories, which facilitates future protocol design and optimization.

We next discuss the detailed design of the MR hardware platform and multi-agent reasoning framework. First, the MR hardware, APEX system integrates state-of-the-art MR goggles, capable of capturing visual, positional, and environmental data from the cleanroom and synchronizing them with the active fabrication workflow. We use MR goggles here as the interactive window between humans and APEX's multi-agent system. Specifically, we programmed the MR goggles with the following capabilities. It can deliver 8K resolution and a 98°–110° field of view with only 32 ms latency, supporting users to get real-time fabrication images without missing fabrication details (**Fig. 2a** and **Supplementary Table 1, 2**). On top of the rendered environment, APEX system also renders real-time adaptive 3D overlays in the MR workspace (**Extended Data Fig. 1**), displaying feedback to the human, including live parameters, progress indicators, and context-specific alerts. Then, in contrast to conventional augmented reality glasses that usually support only the video input modality[42,43], MR goggles were designed to not only capture live video as input to the AI agent but also save corresponding multimodal information including hand and eye tracking data. These data subsequently provide nuanced descriptions about the user behavior that further assist the agent's analysis and understanding of the fabrication (**Extended Data Fig. 2a-** **d**). Moreover, we equipped MR goggles with simultaneous localization and mapping algorithm (SLAM)[44], which provides a reconstructed 3D map of each cleanroom or laboratory with real-time human localization for APEX system (**Extended Data Fig. 2e, f**). Using a spatially resolved, lab-specific 3D map and human location tracking, APEX system understands the interaction between the human and the physical layout of the experiment environment (e.g., associating the user's hand and eye movements relative to instruments and samples to specific experimental action) and enabling it to reason and recognize fabrication steps with higher relevancy and accuracy.

Second, in the multi-agent reasoning framework, APEX system couples reasoning with physical execution by using four core multimodal LLM-assisted agents—Planning, Step-tracking, Context, and Analysis agents—that when orchestrated, continuously understands and evolves during complex fabrications (**Fig. 2b**).

Specifically, the Planning agent first generates high-level device fabrication protocols, incorporating domain knowledge and previous fabrication experiences in its knowledge base. It then produces and sequences SOPs of individual fabrication steps in the protocol and derives per-SOP experiment and step-tracking plans to control the workflow of other agents. The Planning agent's workflow dynamically switches based on the user intent and the maintained SOP atlas. The agent uses chain-of-thought reasoning to determine whether existing SOPs in the atlas can compose a protocol to match the user intent. In addition, when a new SOP is required to adapt to new fabrication settings, the Planning agent references online literature and documents with search and text generation tools to expand its SOP atlas. With the generated SOPs, we apply in-context learning, prompting the Planning agent with examples to generate experiment and step-tracking plans for each SOP.

The Context agent serves as a vision grounding module that links real-world fabrication data to structured knowledge from the experiment plan, including key equipment, materials, and SOP-defined steps (**Methods**). Through in-context learning, the agent converts each visual frame into a structured representation, containing visible instruments, materials, environmental states, and user actions. These representations form the foundation for downstream reasoning, enabling the Step-tracking agent to track fabrication progress and the Analysis agent to perform error correction, provide next-step guidance, and ensure traceability.

The Step-tracking agent aligns the current visual frame with steps in the SOP by aggregating information in its evolving memory (**Methods**). Its workflow is guided by the step-tracking plan from the Planning agent, which, based on different SOPs, dynamically adjusts the interval at which the agent updates its memory and makes a new step prediction. Importantly, we designed the multimodal memory of Step-tracking agent to maintain structured information of recent frames and evolve temporally, removing and adding frame information as dictated by the step-tracking plan. The Step-tracking agent can therefore aggregate recent frame information in its memory, together with previous frame information provided by the Analysis agent, to conduct robust step prediction of the current frame.

The Analysis agent maintains an evolving memory of the fabrication procedure that facilitates procedural understanding of the experiment and provides various real-time feedback to humans. This memory evolves by integrating and analyzing new frame information with the predicted step, forming a comprehensive experiment history of the current SOP that enables three main functions. First, it provides previous frame information to the Step-tracking agent. Second, it enables the Analysis agent to perform human-in-the-loop interactions with the user by answering user queries about previous operations, detecting procedural errors via analysis of the current and previous frame information, and providing instant error corrections and next action guidance (**Methods**). Third, it provides fabrication experiences for the Planning agent to analyze and improve experiment protocol design.

We systematically evaluated the performance of APEX system multi-agent framework through representative fabrication steps (**Methods**). First, we benchmarked APEX system on its ability to understand fabrication tools. Specifically, we compared APEX system with leading multimodal LLMs, including GPT-5-nano, GPT-4o, Gemini-2.5-flash, and Gemini-2.5-pro in describing MR-captured fabrication tools, evaluating whether the target tool/instrument was generated as part of its description (**Fig. 2c**). Across all cases, APEX system consistently outperformed other multimodal LLMs. For instance, the ability of APEX system to recognize the ultrasonic cleaner in the wafer solvent cleaning procedure outperformed other multimodal LLMs by 36%, with an average improvement of 23% across all instruments. To further quantify APEX system's scene understanding capability, we assessed APEX system's ability to comprehensively and accurately describe and annotate MR frames (**Extended Data Fig. 3a–e** and **Supplementary Note 1**). Across different fabrication contexts, APEX system demonstrated the ability to decompose complex visual information into structured representations of tools, environment and user actions, enabling more actionable reasoning and adaptive guidance. Second, APEX system involves a multi-agent reasoning framework that robustly tracks the entire experiment process, which general multimodal LLMs fail because they lack both the background information of fabrication as well as the ability to understand long context information. To equip APEX system with such abilities, we designed a framework that uses the Planning and Context agents to form a static memory of basic experiment setup context and uses the evolving dynamic context specific to every experiment trial.

We then assessed APEX system's ability to understand human operations throughout extended experimental sequences. In three representative examples, APEX system successfully tracked full procedures with consistently high self-reported confidence across multiple representative SOPs (**Fig. 2d** and **Extended Data Fig. 4**). During the eight-step reactive-ion etching (RIE) process, APEX system localized each frame to the correct procedural step (**Fig. 2e**) with high accuracy, while general multimodal LLMs such as GPT-4o and GPT-5-nano frequently misidentified steps and failed to recognize the final pump-down stage. On average, APEX system achieves 51% higher step-tracking accuracy than existing multimodal LLMs. We further assessed how each part of APEX system contributes to the performance (**Extended Data Fig. 5**). We benchmarked the full APEX system against APEX system with specific components ablated. The APEX system accuracy in step-tracking outperforms the ablated version with only Planning/Context agents, Planning/Context/Step-tracking agents, and Planning/Context/Step-tracking/Analysis agents by 41%, 28%, and 10%, respectively. These results collectively demonstrate that APEX system not only recognizes tools and instruments within diverse fabrication settings but also preserves procedural logic and contextual dependencies across long, multi-step workflows, establishing its robustness in dynamic physical experimentation environments.

**APEX system transfers fabrication experience to novices in the cleanroom**

We first asked how APEX system can support novices in the cleanroom. Novices typically face the following challenge: mastering microfabrication skills traditionally requires months to years of mentorship, as new users must learn to navigate complex, multi-step processes that demand not only procedural accuracy (e.g., correct execution of steps in ordered and accurate parameter selection) but also contextual understanding (e.g., ability to interpret equipment states, time dependencies, and the rationale behind each operation within the overall fabrication workflow).

To address this bottleneck, APEX system augments static SOPs with expert-anchored, context-rich demonstrations that help make tacit execution knowledge explicit. Specifically, during expert fabrication we collect multimodal behavioral recordings—including egocentric video and, optionally hand- and eye-gaze trajectories—and align these recordings to each SOP step (**Methods**). This step-level alignment enables a structured characterization of expert practice, including how operations are executed, when transitions occur, and which physical cues are relevant, thereby supporting both procedural accuracy and contextual understanding. When assisting novices, APEX system uses its multi-agent reasoning framework to interpret the user's ongoing actions against the evolving SOP state (i.e., the dynamically tracked execution state of the procedure), infer the intended next step, and deliver stepwise, real-time guidance that is consistent with both the SOP text and the corresponding expert behavior extracted from the recorded demonstrations. By continuously grounding guidance in the physical context and validating progress step-by-step, APEX systematically captures, transfers, and iteratively refines practical knowledge, enabling consistent and efficient training across fabrication tasks. Conceptually, this mechanism allows APEX system-assisted beginners to rapidly approach the performance of experienced researchers, compressing the timescale of skill acquisition from months or years to hours (**Fig. 3a**).

To confirm the fabrication experience transfer efficiency of APEX system, we performed a series of common microfabrication experiments among experts, novices with APEX system and novices without APEX system, using only operating manuals. First, we compared the behavior of different user groups performing the task of RIE. With only guidance from the device manual, the inexperienced users often failed to complete the full sequence of steps (**Fig. 3b**). They lack real-time step-by-step instructions, and therefore often hesitated, made mistakes or carried out steps in the wrong order. In contrast, in the full operation including sample loading, chamber pumping, parameter adjustment, plasma operation, and chamber venting, the inexperienced users with APEX system were as productive as a typical experienced user (**Fig. 3c, d**). With real-time, stepwise visual cues from APEX system that were aligned with the user's movements (**Extended Data Fig. 6a**), the inexperienced users understood the experiment procedure with higher efficiency and accuracy, leading to expert-level performance. Second, we analyzed the novices' operations in the spin-coating procedure. Cooperating with APEX system, not only can the inexperienced researchers finish the full procedure (**Fig. 3e**), but they can also produce the film thickness within the target range at the first spin-coating (**Fig. 3f**). As a contrast, novices were unable to complete the full procedure using the operating manual alone. They either failed to produce a continuous film or generated films with thicknesses far outside the target range due to operational errors. Throughout the spin-coating process, APEX system continuously interpreted user actions in the cleanroom and provided stepwise feedback from parameter configuration to pre-baking (**Extended Data Fig. 6b**). In a representative spin-coating process, APEX system successfully provided accurate, real-time assistance to a novice user, including guidance on upcoming actions, error correction, and confirmation of ongoing steps. Furthermore, we benchmarked APEX system against generic LLMs on the spin-coating process based on expert rating. On average, APEX system outperformed generic LLMs in both guidance completeness and accuracy, achieving improvements of 30% and 34%, respectively (**Extended Data Fig. 6c** and **Methods**). Third, we performed a standard photolithography pattern test (**Extended Data Fig. 7**). With APEX system guidance, inexperienced users received timely, stepwise feedback and were able to complete the patterning process, producing well-defined patterns comparable to those fabricated by experts. Photographic and bright-field microscopy images revealed precise, defect-free interconnect and electrode patterns across device arrays on the silicon substrates. In contrast, unassisted novice users either failed to complete the procedure or produced distorted features consistent with under- or overdevelopment of the pattern.

Together, these results demonstrate that APEX system enables novices to achieve expert-level fabrication performance through multimodal perception, adaptive feedback, and shared experiential learning, establishing a scalable foundation for reproducible, data-driven, and human–AI collaborative scientific manufacturing.

**Error correction and autonomous traceability**

Next, we examined how APEX system could enhance microfabrication performance for experienced users. Even experienced users in the cleanroom often require many forms of assistance. For example, during simultaneous multi-batch fabrication across different projects, experienced users sometimes still confuse parameters or overlook specific steps. Real-time error detection and correction are therefore essential to prevent such mistakes from compromising the entire device fabrication process. Moreover, for complex, multi-batch fabrication workflows, experienced users can benefit from automated summarization and documentation of the entire fabrication procedure, which supports procedural consistency, improves traceability and integrity of experimental records across batches and projects. To address both real-time error prevention and reliable experiment documentation in complex fabrication workflows, building upon the accurate understanding provided by the multi-agent reasoning framework, APEX system continuously interprets user operations in the context of the evolving SOP. By a comprehensive comparison between the physical operations and guidance from the SOP, APEX system identifies errors in real-time and proposes possible solutions to the user, while simultaneously generating structured, context-aware records of the fabrication process.

Equipped with this pipeline, we next asked whether APEX system can improve fabrication performance for experienced users. First, we showed that APEX system helps human researchers correct human errors in real-time during fabrication (**Fig. 3g, h**). As an example, in the parameter setting step of RIE, a radio frequency setting of 50 W applied for 30 s is required in a representative device batch. APEX system successfully detected deviations from the SOP, identifying the wrong entry of 100 W for 10 s and outputted an alert "*The current settings are incorrect…required 30 s and 50 W"* for immediate correction. APEX system uses its multi-agent architecture to detect action deviations and unsafe user behaviors. To evaluate this capability, we consider four types of errors found in the RIE procedure: (1) premature opening of the chamber before the target vacuum conditions were reached, (2) setting wrong etching parameters, (3) initiation of etching process prior to completion of the required setup, and (4) early termination of the etching process (**Extended Data Fig. 8** and **Methods**). Chosen for their criticality in the RIE procedure, these errors, if uncorrected, can irreversibly affect the outcome of experiment. For error detection, an intuitive approach is to directly prompt the LLM with the MR frame. However, we show that this information alone is insufficient, as the full APEX system exceeds MR frame-prompted LLMs by 51%, 60%, 29%, and 26% on average across the four categories in accuracy. Successive incorporation of the Context agent and the Analysis agent led to gradual improvement in error detection accuracy, with the full APEX system exceeding the ablated version with only the Planning/Context agents and Planning/Context/Analysis agents by an average of 28% and 22%, respectively. We verify that the complete APEX system achieves successful real-time recognition and correction of action deviation and unsafe interactions with an average accuracy of 96% across the four representative failure modes. Second, we further demonstrated that the equipment errors can be identified in real-time, analyzed and corrected (**Extended Data Fig. 9a**). APEX system autonomously collects previously identified errors, storing them in a representation that integrates operator actions and equipment states, enabling the retrieval of expert solutions when similar errors are encountered in future operations.

Importantly, the same step-level, context-aware understanding that enables real-time error detection and correction also provides a foundation for persistent recording and retrospective analysis of fabrication processes. We further presented that APEX system can autonomously generate structured experimental summaries containing step identifiers, parameters and environmental snapshots (**Extended Data Fig. 9b, c**). In the photolithography process, the APEX system precisely captured timestamped, important readings from the maskless aligner equipment "*Expose time: ~6.2 s*", "*Load indicator: Green On*", preserving experiment details for analysis, user queries and reflection. As an example, in the RIE step, the human user directly queries the APEX system's generated experimental summary and receives responses that reference specific experiment parameter "*You set the timer for 30 seconds.*" and SOP step "*You set the timer during step 4. At timestamp [01:21], you began inputting the time…* ". These results show that APEX system enables human reflection and analysis through accurate experiment documentation.

Overall, APEX system (1) correctly identified and assisted in real-time correction of procedural errors, and (2) autonomously generated comprehensive experimental documentation encompassing all lithography, deposition, etching processes. Through this bidirectional interaction, APEX system transformed fabrication from an isolated manual endeavor into an interpretable, collaborative process. Rather than replacing human expertise, it amplifies precision, consistency, and reproducibility. Together, these findings demonstrate that APEX system enables continuous operation across diverse fabrication stages and supports the successful completion of complex, multi-step scientific manufacturing in collaboration with human researchers.

**APEX system co-develops new protocols with researchers**

A core challenge in micro- and nanofabrication is the development of stable fabrication protocols under intrinsic information incompleteness. Protocol decisions must be specified prior to fabrication, whereas critical knowledge such as material behavior, interfacial stability, and stress evolution only emerges during physical execution. As a result, fabrication outcomes are shaped by tightly coupled effects of process parameters, step ordering, and execution-dependent phenomena that cannot be reliably predicted. Existing approaches provide limited remedies. Protocols reported in literature often exhibit limited transferability across different materials, steps, specific tools and facilities. While digital tools, including LLMs, can retrieve and summarize fabrication knowledge, they remain detached from physical execution and therefore cannot revise tool-wise protocols in real-time based on intermediate outcomes. Consequently, protocol development continues to rely on prolonged empirical trial-and-error by human researchers.

Addressing this challenge requires a system to continuously integrate observations of physical experiment into protocol development. APEX system bridges this gap in a closed-loop framework by coupling two critical processes: (1) protocol design and (2) physical fabrications. Core to protocol co-optimization is the reflection mechanism that incorporates prior knowledge and real-time feedback. First, prior to fabrication, APEX system generates a high-level protocol aligned to user intent. Using the SOP atlas, APEX system translates the protocol to detailed steps by either customizing or creating new SOPs such that all are compatible with the specific materials and tools. The Planning agent in APEX system iterates on the detailed steps generated to ensure output is compatible with previous protocol designs, fabrication records in the knowledge base, and feedback provided by human researchers. As an example, APEX system learns to incorporate data found in the fabrication facility and designs a location-specific protocol through derivation of parameters. Second, through its integration in the physical environment, APEX system incorporates observations from fabrication into its protocol generation process, allowing it to reason and generate possible causes of the experiment failure and improvements to the SOPs. APEX system continuously incorporates information collected via its natural presence in physical space, allowing the once rigid protocol design process to rapidly evolve in response to real fabrication outcomes.

To verify performance of the defined pipeline, we first assessed whether APEX system could enable parametric co-optimization, addressing the long-standing difficulty of extensive parameter sweeps across tools and recipes. We demonstrated that APEX system enables guided and constrained exploration of the spin-coating parameter space by incorporating measured film thickness data (**Extended Data Fig. 10a** and **Methods**). Relative to exhaustive parameter sweeps, this strategy substantially reduces the effective search space and accurately predicts the spin-coating speed and styrene–ethylene–butylene–styrene (SEBS) concentration parameter space that yield target SEBS thicknesses, achieving the targeted parameter identification with only one-sixth of the experimental effort (2 measurements with APEX system versus 12 measurements without APEX system). Second, we proved that APEX system can identify incompatibility among materials, solvents, and processing steps and suggested alternative processing routes to mitigate these effects (**Extended Data Fig. 10b**). Soft SEBS in Remover PG is identified by APEX system as non-compatible and APEX system immediately proposed a non-solvent method for gold patterning with SEBS substrate.

We then asked whether APEX system can co-develop a new, stable protocol that originally requires expert years to develop, co-fabricate and analyze devices together with researchers. We determined to examine a fundamental bottleneck in microfabrication arising from the incompatibility between soft elastomers and conventional fabrication processes. Elastomers are well recognized as brain-level soft encapsulation materials for minimizing mechanical mismatch at the neural interface and enabling stable single-unit tracking; however, they are notoriously incompatible with conventional microfabrication processes due to solvent-induced swelling, sensitivity to thermal history, and susceptibility to mechanical deformation. Most elastomers cannot be fabricated into ultrathin planar neural probes, and prior successes have been limited either to fluorinated elastomers specifically engineered for microfabrication compatibility[45] or to thick architectures formed by rolling a single elastomer layer into structures several hundred micrometers in diameter, which inevitably increase flexural rigidity[46]. As a result, developing fabrication protocols that can be tailored to different elastomers for ultrathin, soft neural probes is highly significant but exceptionally challenging.

We validated the co-development pipeline by designing a protocol with APEX system for a-few-micrometer-thick, ultrasoft elastomeric neural probes using SEBS elastomer as a demonstration (**Fig. 4a, b**). APEX system first generated a customized fabrication protocol by recognizing that directly spin-coating SEBS as the top encapsulation layer would dissolve the underlying SEBS layer. To avoid this issue, APEX system selected a release–float–transfer approach to form the top SEBS encapsulation (**Methods, Supplementary Fig. 1-4** and **Supplementary Note 2**). During fabrication, APEX system found that the free-standing SEBS top layer was difficult to transfer onto the device wafer; in response, APEX system immediately revised the workflow and adopted a polydimethylsiloxane (PDMS)-mediated transfer strategy. This APEX system-assisted process enables wafer-scale realization of ultrathin, planar SEBS neural probes (**Fig. 4c–f**). Subsequent implantation and *in vivo* neural recordings verified that the ultrasoft neural probe enables stable, single-unit–resolution neural signal recording (**Fig. 4g–i**). Importantly, SEBS serves here as a representative elastomer rather than a material-specific solution. By operating at the level of execution-aware protocol co-development, APEX system can generalize to a broad class of elastomeric materials that are traditionally incompatible with microfabrication, enabling device realizations that are designable in principle but previously unrealizable in practice.

Together, these results show that APEX system can identify and resolve tightly coupled process constraints by directly linking execution-dependent observations with protocol reasoning. This closed-loop, execution-aware approach substantially shortens protocol development time and enables fabrication outcomes that are difficult or impractical to achieve using existing methods that operate exclusively in either the digital or physical domain.

## Discussion

APEX system, as a demonstration of human–AI co-embodied system that integrates human operators, agentic AI, and a wearable MR interface into a closed-loop system, enabling continual perception of ongoing procedures, long-horizon reasoning over operation workflows, and real-time interaction with humans. During experiment and manufacturing, APEX system functions as an active collaborator, anticipating upcoming steps, detecting deviations, and delivering real-time guidance adaptively. Furthermore, APEX system generates structured experimental records that not only transform into traceable and reproducible knowledge but also enable *post hoc* analysis and co-optimization of protocols with humans.

Looking forward, the APEX system establishes a foundation for collaborative experimental and microfabrication workflows in which human and AI operators jointly analyze experiments and co-develop protocols during execution. We illustrated the co-development capability through the fabrication of a-few-µm-thick, tissue-level soft SEBS-based neural probes. Although brain-level soft elastomeric materials such as SEBS are attractive for reducing mechanical mismatch and thus enabling stable single neuron tracking, they are generally incompatible with standard microfabrication workflows. By tracking fabrication steps in real-time and adapting encapsulation and transfer strategies, APEX enabled a new, generalizable protocol for wafer-scale fabrication of planar, a-few-µm-thick SEBS neural probes, yielding stable single-neuron recordings within one week—an outcome previously impractical using conventional approaches.

Despite these advances, challenges remain. Firstly, APEX system incorporates domain knowledge through a context window, which has restricted capacity and limits APEX system's ability to incorporate the ever-growing knowledge base constructed from both local fabrication experiences and online search. Fine-tuning of LLMs used in APEX system using the knowledge base could enable APEX system to internalize domain knowledge more efficiently and therefore generalize to a wide range of fabrication settings. Retrieval-augmented memory could serve as another option to help APEX system filter out the most relevant information to prioritize. Second, APEX system currently obtains information about the lab through MR goggle perception. Although it provides accurate information, it requires the user to interact with the devices regularly to check the real-time status and may interrupt the experiment workflow. Future work could mitigate this challenge by building an Internet-of-Things network in the cleanroom, connecting both APEX system and the cleanroom instruments to the same network. With the cleanroom instruments periodically publishing their real-time status, APEX system automatically obtains real-time information streaming for agentic reasoning without human intervention. Finally, current cloud-hosted LLM inference adds to the reasoning delay and reduces data security of APEX system. On-device deployment of multimodal LLMs on MR goggles enables local LLM calls during experiment and manufacturing, reducing communication delay and enhancing both data security and connection stability.

## Methods

### 1. APEX structure and benchmarks

#### The Planning agent develops new protocols and generates comprehensive tool-specific fabrication SOPs

The Planning agent synthesizes user intent into executable, specific, and ordered lists of steps executed in the cleanroom. First, before entering the cleanroom, the user develops a protocol with the Planning agent by iteratively prompting the agent with their fabrication requirements, optionally providing existing protocols from literature and images from the experiment setup to iteratively compose a document including the ordered phases (e.g., ultraviolet (UV) flood exposure, dextran sacrificial layer) in the cleanroom fabrication protocol. Next, the Planning agent decomposes each phase in the fabrication protocol into one or more ordered SOPs, each corresponding to one process in the cleanroom (e.g., spin-coating) and comprises a list of ordered steps with specific parameters tailored to the experiment requirements, experiment setting, and the user intent. In the decomposition of fabrication phases, we equip the Planning agent with the SOP atlas, a repository of expert-curated SOPs that could guide in the completion of individual atomic workflows (**Supplementary Table 3**), empirical knowledge summarized by APEX system from previous fabrications (e.g., devices containing polydimethylsiloxane are typically baked at approximately 70 °C), and the latest status of the specific cleanroom, such as that the h-line dose rate of the contact mask aligner at the Harvard University Center for Nanoscale Systems is observed to be 89 $mJ\ cm^{-2}\ s^{-1}$.

At the beginning of each procedure, the Planning agent generates the experiment plan, which contains equipment/material used in the matching SOP and its ordered list of steps, and the step-tracking plan, which includes the step-tracking memory size and the step prediction interval (**Supplementary Table 4**), both in units of MR experiment frames, and the same list of ordered steps. The step-tracking memory size sets the size of the Step-tracking agent's evolving short-term memory. The prediction interval defines the number of MR experiment frames between consecutive step predictions. The Planning agent uses Gemini-3-pro-preview (temperature = 0) for its LLM backend.

#### Context agent aligns visual information to experiment plan

The Context agent contextualizes the visual information in each MR frame under the setting of the current SOP. Provided with the experiment plan and a MR frame, the Context agent generates the scene context, a list of equipment/material that appears in the frame, a description of the environment, and a summary of user actions. Crucially, the Context agent is limited to not inferring the intentions of the user in the MR frame (**Supplementary Table 5**). While the intentions of the user are important for scene understanding (e.g., whether the user is retrieving or inserting a wafer into the RIE machine), this task is restricted to downstream processes in APEX system after the user's step progression in the current SOP has been confirmed by the Step-tracking agent. Moreover, the separation of scene understanding enables the Context agent to operate completely independently (after the experiment and step-tracking plan have been generated). The deployment of the Context agent and the generation of scene context occur at the rate of the headset's MR frame output, thereby allowing near-real-time operation of APEX system. The Context agent uses Gemini-2.5-flash-preview-09-2025 (temperature = 0) for its LLM backend.

**The Step-tracking agent localizes user actions in standard operating procedures**

The Step-tracking agent measures user progress in the current scene context in the SOP at the interval of the step-prediction interval and determines the step number in the SOP based on its two memories. The short-term memory of the Step-tracking agent is a continuous set of past scene contexts kept at the size set in the step-tracking plan and is designed to capture the transient dynamics of rapid user actions, for example, opening a bottle of photoresist. The long-term memory comprises the complete, time-stamped output of the Analysis agent accumulated since the beginning of the SOP and is used to inform the Step-tracking agent of the user's overall progression. The Step-tracking agent leverages both memories and the current scene context to conduct a reasoning process prior to localizing the user's current context in the SOP (**Supplementary Table 6**). First, it evaluates the previously predicted step by identifying three required sub-actions (the full completions signify that the step is complete) and then assessing the presence or absence of evidence for each sub-action. Second, if the previously predicted step is determined to be complete, the agent confirms advancement to the step corresponding to the current scene context; otherwise, it retains the previously predicted step. The Step-tracking agent outputs its decisions as a list of three most probably SOP steps and the self-reported confidence value for each, ranging continuously from zero to one, where zero denotes "very weak evidence" and one denotes "very strong evidence". Under ideal conditions, the previously predicted step and the long-term memory at the time of step prediction correspond to the scene context observed exactly one step-prediction interval prior to the current MR frame. In practice, to support near-real-time operation in the cleanroom, the Step-tracking agent does not enforce full temporal synchronization of all memory updates at each prediction cycle. Instead, at each step-prediction interval, the agent waits only for completeness of the short-term memory before generating a prediction. The Step-tracking agent uses Gemini-2.5-flash-preview-09-2025 (temperature = 0) for its LLM backend.

**The Analysis agent enables step-aware scene analysis and co-optimizes new protocols**

The Analysis agent aggregates findings from the previous agents into a shared, consolidated report. After the step prediction, the Analysis agent reasons and contextualizes the scene context against the predicted step (**Supplementary Table 7**). Specifically, it maps the action performed by the user in the scene context to a specific sub-action of the predicted step. Then, the Analysis agent contextualizes the observed behavior and selectively records equipment states and readings that are most relevant to the current step. Importantly, by leveraging step-level information provided by the Step-tracking agent, the Analysis agent transforms the objectively recorded scene context into step-aware representations that capture both the operational significance of the user's actions and their relevance within the broader procedural workflow. The Analysis agent further analyzes and converses with the user offline for parameter and procedural optimization of the generated SOPs from the Planning agent. The Analysis agent uses Gemini-2.5-flash-preview-09-2025 (temperature = 0) for its LLM backend.

**Human-in-the-loop interactions**

The Step-tracking and Analysis agents jointly orchestrate APEX system–human interaction under four scenarios: (1) protocol co-optimization, (2) human query and reflection, (3) error detection and correction, and (4) experiment guidance. During protocol co-optimization, the user collaborates with APEX system via the Planning agent to develop the fabrication protocol, after which the resulting set of SOPs is broadcast downstream for use by other agents. At any point during fabrication, the user may query the Analysis agent to confirm procedural progress; such queries are answered with evidence grounded in the agent's accumulated long-term memory. In addition, the Analysis agent provides error detection and corrective guidance at each confirmed step. Specifically, using a carefully designed prompt (**Supplementary Table 8**), the Analysis agent integrates step predictions from the Step-tracking agent with observed actions recorded by the Context agent, compares expected and observed user actions, and reports detected errors or potentially unsafe behaviors, a summary of the user's current actions, and anticipated next steps.

**Performance assessment of APEX system**

The performance assessment of APEX system was conducted across three of its core functions in the context of cleanroom fabrication: perception, reasoning, and action.

The APEX system perception function was validated against two datasets (**Supplementary Note 3**). In the scene context dataset, we selected MR frames across five representative fabrication procedures. For each MR frame, APEX system's generated scene context was scored by a microfabrication expert on the accuracy and completeness of its response. Results per fabrication procedure were reported as the average of its scores against the prepared rubric (**Supplementary Note 1**). Next, APEX was tested against generic LLMs. In the equipment/material dataset, a microfabrication expert curated images of six commonly used pieces of equipment and material in the Harvard Center for Nanoscale Engineering under varied lighting and operating speed conditions. For each image, APEX system and state-of-the-art multimodal LLMs were tested to predict the presence of the target material or equipment across six independent trials; results were compared against the ground truth using accuracy.

The reasoning function of APEX system was evaluated by testing its step-tracking ability (**Supplementary Note 4**). The step-tracking dataset comprised of three representative complex fabrication processes (reactive-ion etching, spin-coating, and photolithography), containing MR frames spanning every step in the processes. First, APEX system and its ablated versions were tested to predict the correct step among the complete set of steps in the corresponding SOP in three independent trials. The ablated versions include Planning/Context agent (predicting step with step-tracking plan, experiment plan and current context), Planning/Context/Step-tracking agent (predicting step with step-tracking plan, experiment plan, and current context, and the structured frame information in the memory of the Step-tracking agent), Planning/Context/Analysis agent (predicting step with step-tracking plan, experiment plan, current context, and history information in the memory of the Analysis agent). Second, to evaluate the performance of baseline LLMs, they were prompted with the complete set of steps in the SOP and the MR experiment frame to predict the correct step. The overall step prediction and stepwise predictions results were compared against the ground truth using accuracy.

Next, the action function of APEX system was evaluated by assessing its ability to correct user errors and provide accurate feedback (**Supplementary Note 5**). In the error dataset, four representative user errors in the reactive-ion etching process were curated and tested in five independent trials on APEX system and its ablated versions, which include the Planning/Context agent (providing error correction with step-tracking plan, experiment plan and current context), Planning/Context/Analysis agent (providing error correction with step-tracking plan, experiment plan, current context, and history information in the memory of the Analysis agent), and the fully-ablated configuration (providing error correction with the full list of steps in the SOP and the MR experiment frame). Results were compared against the ground truth using accuracy. Moreover, APEX and generic LLMs were tested to generate experiment error alerts, current action and next action guidance for representative MR frames from the spin-coating procedure; results were reported as the average of its scores evaluated by the prepared rubric (**Supplementary Note 1**).

In performance assessment of APEX system, ablated configurations of APEX system, and baseline LLMs that were composed of multiple independent trials, reported accuracies were computed by taking the mean over all individual data points across all trials. Error bars denote the standard error of the mean (s.e.m.), computed over the same set of data points with zero degrees-of-freedom.

## 2. Materials and electronics

### Materials

SEBS (Tuftec H1062 series) was obtained from Asahi Kasei. Photoresists and developers in the nanofabrication procedure were obtained from MicroChem Corporation unless otherwise specified.

### Ultrasoft SEBS neural probe fabrication

The protocol generation process was conducted jointly between user and APEX system (**Supplementary Fig. 1-4** and **Supplementary Note 2**).

### Thickness measurements

Film thickness measurements were performed with a DektakXT contact profilometer (Bruker). Profiling was carried out using a stylus load of 3 mg at a lateral scan rate of 0.67 μm $s^{-1}$, with height data corrected by two-point surface leveling in the native analysis software. For each wafer, thickness measurements were acquired at three predefined locations: the wafer center and two lateral positions located 3 cm to the left and right of the center, respectively.

## 3. Animal experiments

### Animals and ethical compliance

All animal experiments in this study were approved by the Institutional Animal Care and Use Committee (IACUC) of Harvard under animal protocol no. 20-05-368-1 and complied with the National Institutes of Health (NIH) guidelines for the care and use of laboratory animals. Adult C57BL/6 mice (Charles River Laboratories, male, 33 weeks of age) were used throughout this study. Mice were kept on a regular 12 h/12 h light/dark cycle and fed ad libitum food and water and housed at a temperature of 22 ± 1 °C and with humidity ranging from 30% to 70%.

### Stereotaxis implantation

To ensure sterility, all metal tools used during surgery were cleaned with an autoclave before use. During intracranial implantation surgery, mice were anesthetized using 2–3% isoflurane for induction and 0.75–1% for maintenance under anesthesia. The surgery was conducted on each mouse, which was placed on a heating pad on a stereotaxic frame with two ear bars and one nose cone. A 2 × 2 mm craniotomy was performed with a sterile scalpel to expose the cortical surface, followed by removal of the dura mater. Stainless steel screws sterilized with 70% ethanol were inserted into the cerebellum to serve as ground electrodes. Planar SEBS neural probes, attached to a customized tungsten shuttle (50 μm in diameter), were secured to a micromanipulator on a stereotaxic frame. This micromanipulator was used to carefully insert the electronics into the brain to the desired depth. The craniotomy was then secured with a silicone elastomer (World Precision Instruments), and ceramic bone anchor screws and dental methacrylate were used to secure the flat flexible connector and electrode set onto the mouse's skull.

### Electrophysiological data acquisition and spike sorting

For data acquisition, a customized printed circuit board (PCB) was used to connect to the flexible flat cables and Intan head stage. The flexible neural signals were subsequently recorded using the Intan RHD recording system (Intan 1024ch Recording Controller) at a sampling rate of 30 kHz. All recordings were performed under 0.75% isoflurane anesthesia. The setup was placed on an optical table and covered by a Faraday cage. Electrical recording processing and sorting were performed offline using a custom fully automated Python pipeline using MountainSort4[47] (**https://github.com/flatironinstitute/mountainsort**) under the SpikeInterface[48] (**https://spikeinterface.readthedocs.io/en/latest/**) module. In brief, raw recordings were first filtered using a band-pass filter in the 300–3,000 Hz frequency range. Then, common median reference was removed to reduce the common mode by subtracting the median of all electrode channels from each channel. Each mouse's electrophysiology data was sorted individually. Spike waveforms were extracted using a spike detection threshold of 4.5 times the standard deviation away from the mean, depending on the base recording noise level. Manual curation used on mean waveform templates was output from MountainSort4 and putative unit candidates were excluded from analysis if they corresponded to noise.

**Acknowledgments**
J.L. acknowledges the support from NSF through the Harvard University Materials Research Science and Engineering Center Grant No. DMR-2011754; the Harvard University Center for Nanoscale systems supported by the NSF. N.L. acknowledges the support from NSF AI institute 2112085.

**Author contribution**
J.L. and X.L. conceived the idea. X.L., Y.Z., N.L., Y.G. and J.C. developed the APEX system method. L.L., Y.Z., S.W., C.W. and R.Z. developed the MR goggle software customized for APEX system. X.L. and Y.G. took all experiments. Y.Z. and Y.G analyzed the videos and performed benchmarks. H. S., Y. H. and N.L. provided critical insights to the APEX system structure. All authors contributed critical discussions and input on the methods, figures and manuscripts. J.L. supervised the study.

**Competing interests**
J.L. is a cofounder of Axoft, Elastro, and AIScientists, Inc.

**Additional information**
Correspondence and requests for materials should be addressed to Jia Liu.

## Figures and Figure Legends

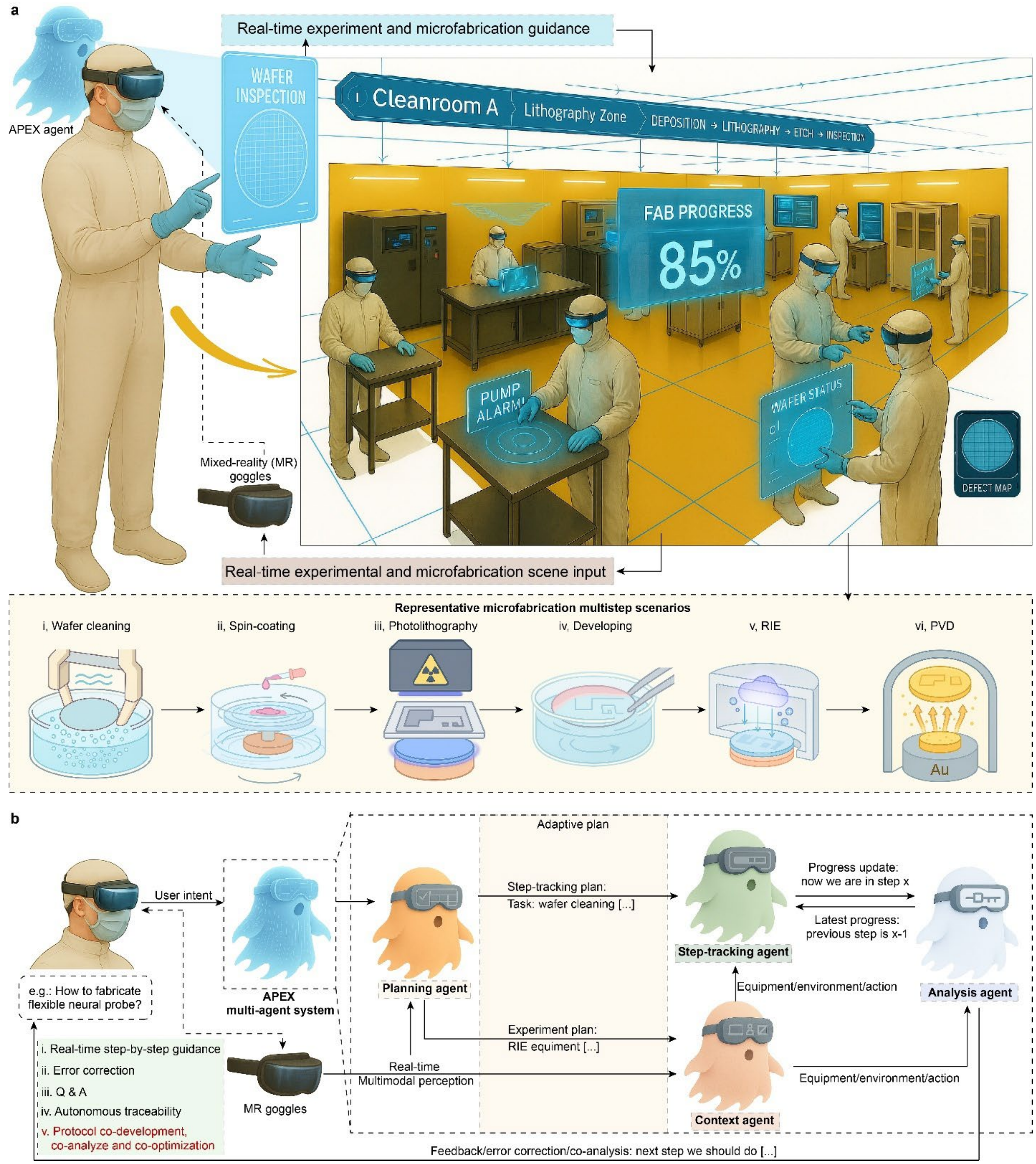

**Fig. 1 | APEX system: agentic-physical artificial intelligence (AI) experimentation system with mixed-reality (MR) for human–AI collaboration in scientific experimentation and intelligent manufacturing. a,** APEX system schematics demonstrating coupling of real-world experimentation and intelligent manufacturing for microfabrication with agentic reasoning. In real-time, APEX system interprets multimodal streams, understands experiments, tracks procedural

progression, stores critical experimental parameters and data, and generates stepwise feedback to researchers. Representative microfabrication scenarios including wafer cleaning, spin-coating, photolithography, developing, reactive ion etching (RIE) and physical vapor deposition (PVD). **b,** Architecture of APEX system showing its four sub-agents: Planning, Step-tracking, Context and Analysis agent. The multi-agent system integrates human physical execution with agentic reasoning, enabling protocol design, task planning, and real-time analysis, documentation, and guidance of human researcher procedures. Human researchers communicate their goals to the Planning agent, which generates detailed protocols for the fabrication, and transmits the designed task sequences, experiment setup to downstream agents. The Context agent interprets the experiment environment jointly with the experiment plan and user action as scene context. The Step-tracking agent leverages the step-tracking plan to align scene context to tasks in the fabrication protocol. The Analysis agent documents and interprets experiment progress, enabling it to deliver feedback to researchers, forming a closed human–AI loop.

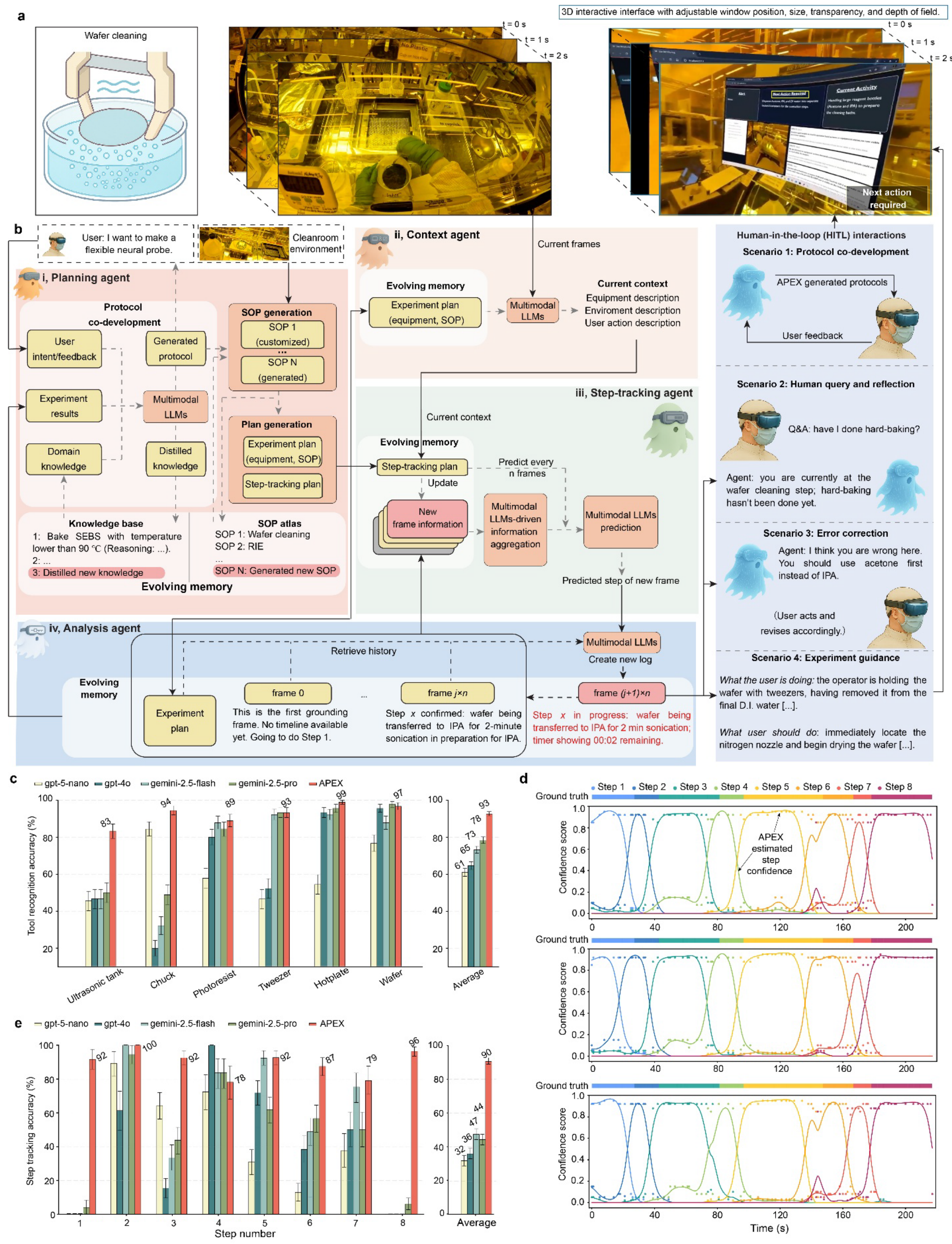

**Fig. 2 | Architecture and performance of the APEX system. a,** Human–AI collaboration for scientific experimentation and intelligent manufacturing scenarios. Example of a wafer cleaning

procedure performed in the cleanroom with APEX system. Egocentric MR video streams and task context are continuously analyzed by APEX system, which interfaces the user with stepwise guidance and corrective alerts using the MR interface. **b,** Detailed multi-agent architecture of APEX system. The Planning agent maintains an evolving standard operating procedure (SOP) atlas and the knowledge base, which, when combined with the user intent, is used to generate an experiment-specific plan/setup and protocol for step-tracking. The Context agent interprets the current scene by linking MR perception (equipment states, user actions, environment) to the current SOP. The Step-tracking agent aligns ongoing human actions with the planned procedure in real-time and evaluates procedural progress. The Analysis agent compiles these updates into a structured, time-stamped experimental log that supports traceability, real-time feedback to the user, and *post hoc* review. Together, these multimodal large language models/vision language models (LLMs/VLMs)-driven agents form a closed human–AI loop that couples agentic reasoning with physical execution. **c,** Bar plots quantifying the device/equipment recognition accuracy of APEX and other LLMs (mean ± s.e.m., $n$ = 90 MR experiment frames input). APEX system achieves higher recognition accuracy than state-of-the-art multimodal LLM baselines (GPT-4o, GPT-5-nano, Gemini-2.5-flash, Gemini-2.5-pro) across diverse cleanroom instruments and fabrication tasks, demonstrating context-aware perception of real manufacturing environments. **d,** Representative temporal traces indicating self-reported APEX step-tracking confidence from three RIE processes. APEX system maintains high self-reported confidence scores as the RIE advances, reflecting stable alignment between human actions and SOP. Each dot represents one MR experiment frame; temporal prediction is visualized by traces smoothed over the dots using radial basis functional interpolation with a thin-plate spline kernel ($\lambda$ = 100). **e,** Bar plots quantifying the SOP step-tracking accuracy of APEX and other multimodal LLMs (mean ± s.e.m., $n \geqslant 18$ MR experiment frames input). In one representative RIE process, APEX system outperforms baseline LLMs in assigning the correct SOP step to each MR experiment frame, achieving higher stepwise and overall accuracy, demonstrating APEX system's ability to understand and monitor procedural progression during fabrication.

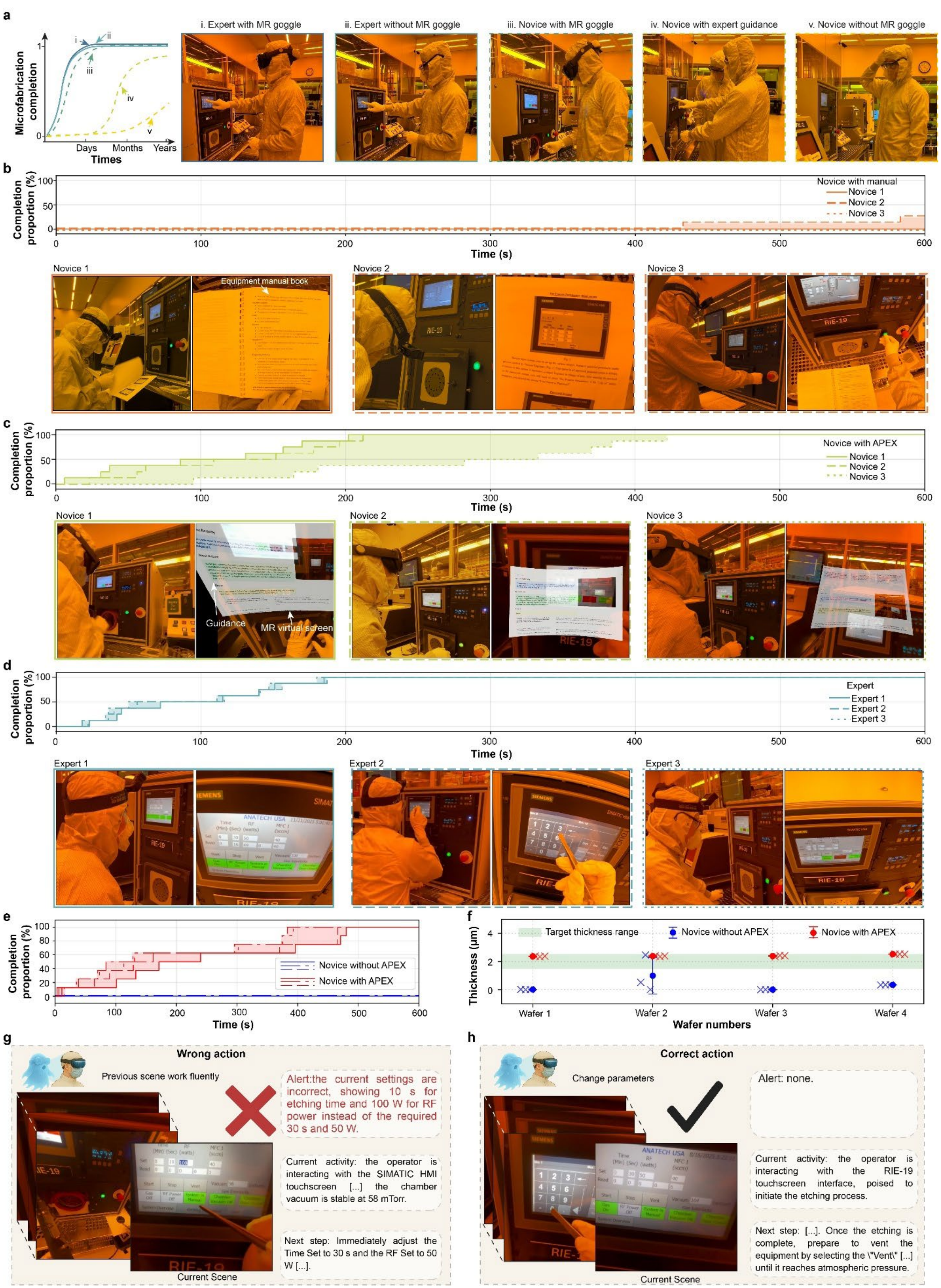

**Fig. 3 | Scalable and transferable scientific experimentation and intelligent manufacturing knowledge enabled by the APEX system. a,** Schematic illustration showing that a novice user reaches performance comparable to that of an expert in the RIE process within a single session under APEX system guidance. **b–d,** RIE process completion proportion as a function of time. Novices only using the equipment manual (**b**) failed to complete or start the process in the time of the expert's completion, while novices using APEX system (**c**) showed expert-level proficiency (**d**) within a single session, demonstrating APEX system's ability to distill and distribute procedural intelligence to users across the full skill spectrum. **e,** Spin-coating process completion proportion as a function of time comparing novice researchers with and without APEX system guidance. In **b–e,** solid and dashed lines indicate disparate experienced and novice researchers; shaded region represents range of maximum and minimum step completion proportion at each time point. **f,** Film thickness of samples spin-coated by novices with and without APEX system guidance. Green shaded region indicates target film thickness in representative spin-coating process; scatter solid dot and error bar indicate trial mean ± s.e.m.; each cross represents an individual film-thickness at different locations on the same wafer ($n$ = 3 measurements). **g–h,** Representative error correction and validation during RIE. APEX system autonomously detects incorrect parameter inputs and provides immediate MR feedback to ensure compliance with SOP-configured settings, preventing procedural drifting and potential fabrication failure. APEX system resumes to monitor procedure executions upon correction.

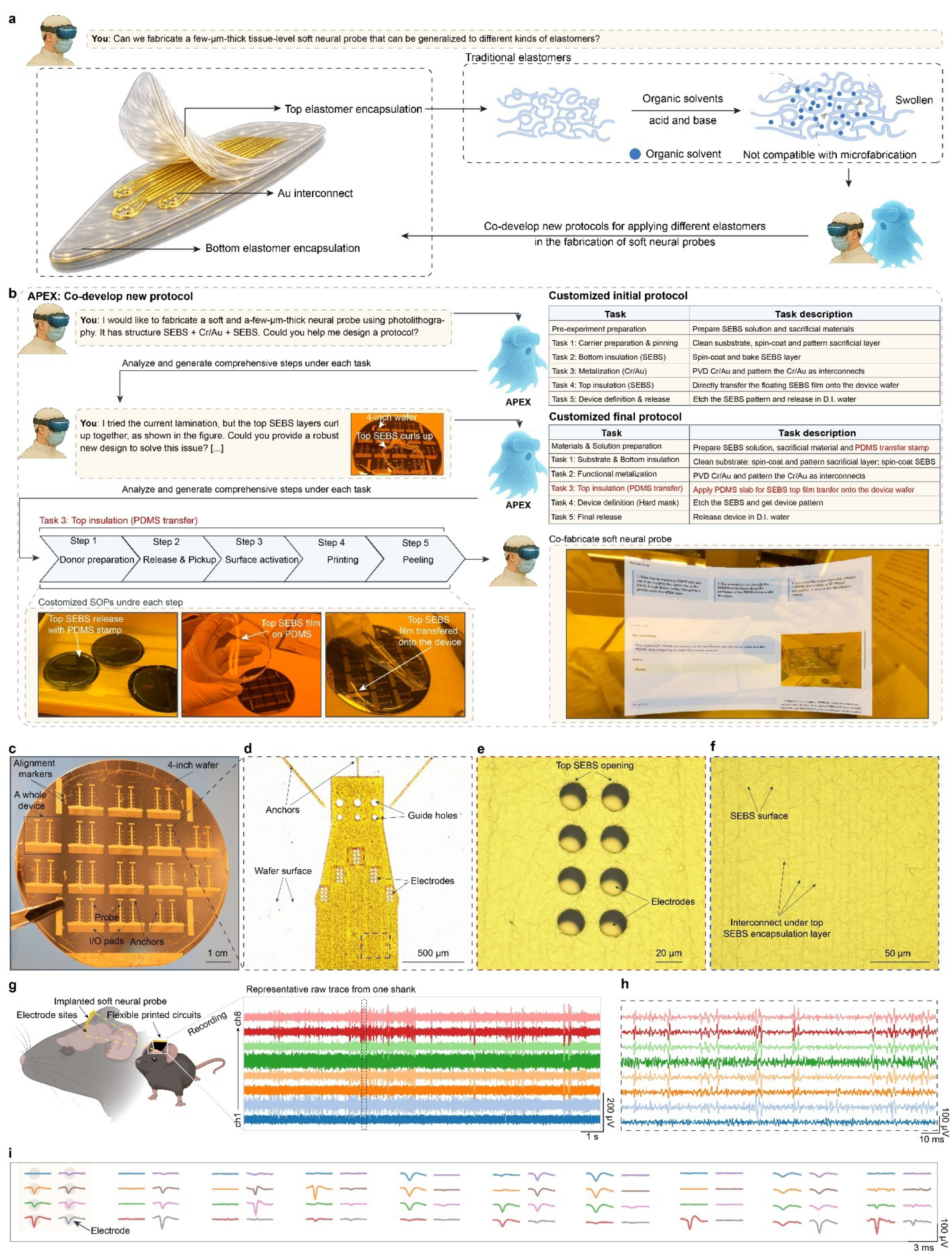

**Fig. 4 | APEX system co-develops and co-optimizes new protocols with researchers. a,** APEX system accelerates the discovery of fabrication protocols for novel application of elastomer in soft

neural probes. **b**, APEX system generates and iteratively refines a fabrication protocol for a-few-micrometer-thick, ultrathin tissue-level soft styrene-ethylene-butylene-styrene (SEBS)-encapsulated neural probes through interactive collaboration with researchers. **c–f,** Representative optical and microscopic images of the SEBS soft neural probe at different length scales, illustrating device layout on a 4-inch wafer (**c**), structural features (**d**), electrode openings (**e**), and encapsulated interconnects (**f**). **g–i**, Representative neural recordings acquired with the APEX system–researcher co-developed ultrathin SEBS soft neural probe implanted in a mouse. Raw voltage traces from multiple channels over a 20 s recording window (**g**) and a corresponding zoomed-in segment (**h**) demonstrate robust neural activity and simultaneous detection of spike events across channels. (**i**), Representative mean single-unit spike waveforms captured concurrently by eight neighboring electrodes, derived from the data presented in **g** and **h**, highlighting the spatially distributed, single-unit resolution recording capability of the probe.

**a**

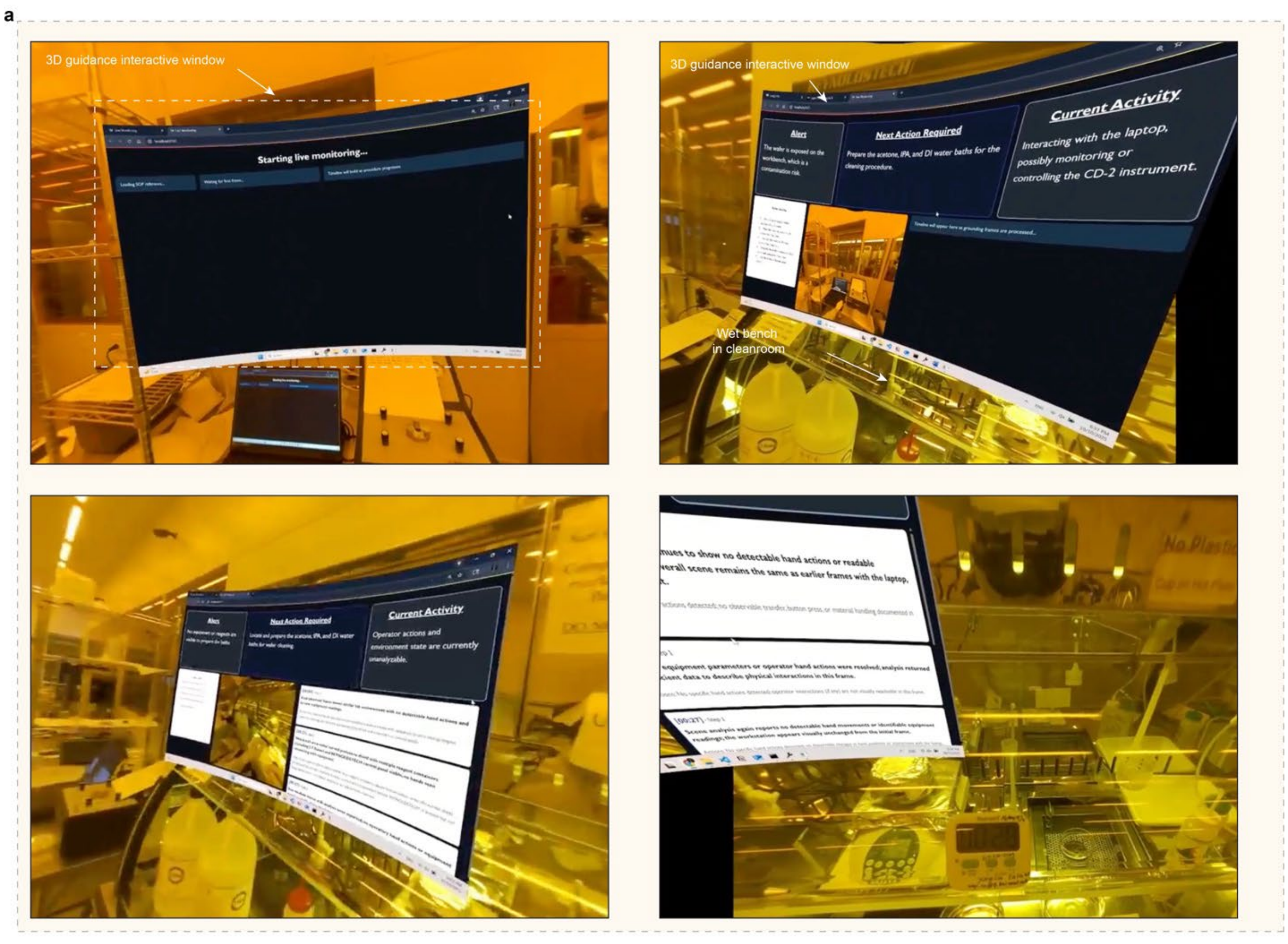

**b**

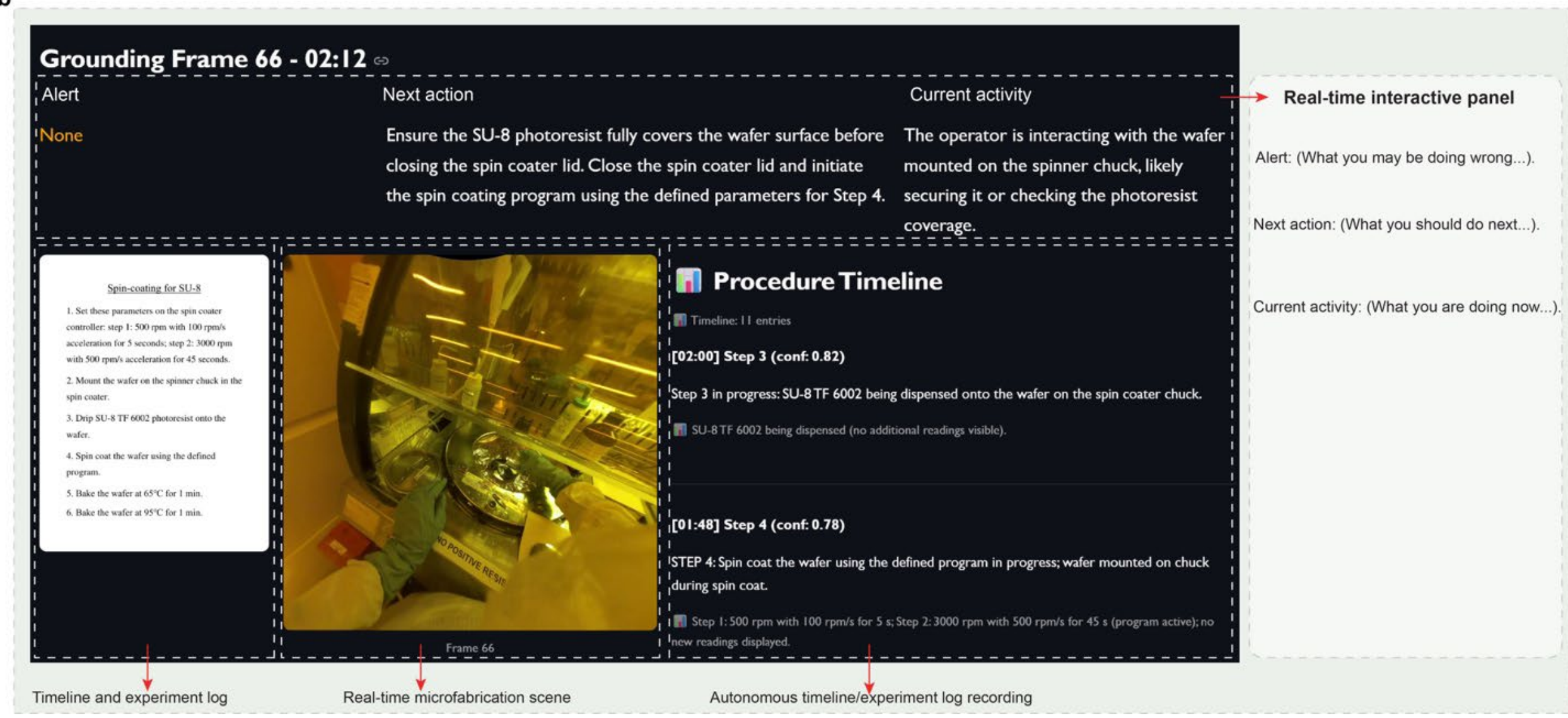

**Extended Data Fig. 1 | Real-time MR interface for human–AI collaboration in cleanroom experimentation and manufacturing. a,** Immersive 3D MR guidance panels displayed in the cleanroom environment. Researchers interact with contextual overlays showing live procedures, progress reports, and adaptive next-step or error prompts synchronized with the ongoing

fabrication scene. **b,** Interactive physical AI interface linking MR visual streams automatically generated experiment timelines. The interface presents contextual alerts, current actions, and predictive guidance in real-time, forming a continuous perception-reasoning-action loop that supports human–AI collaboration during experimentation and manufacturing in cleanroom for microfabrication.

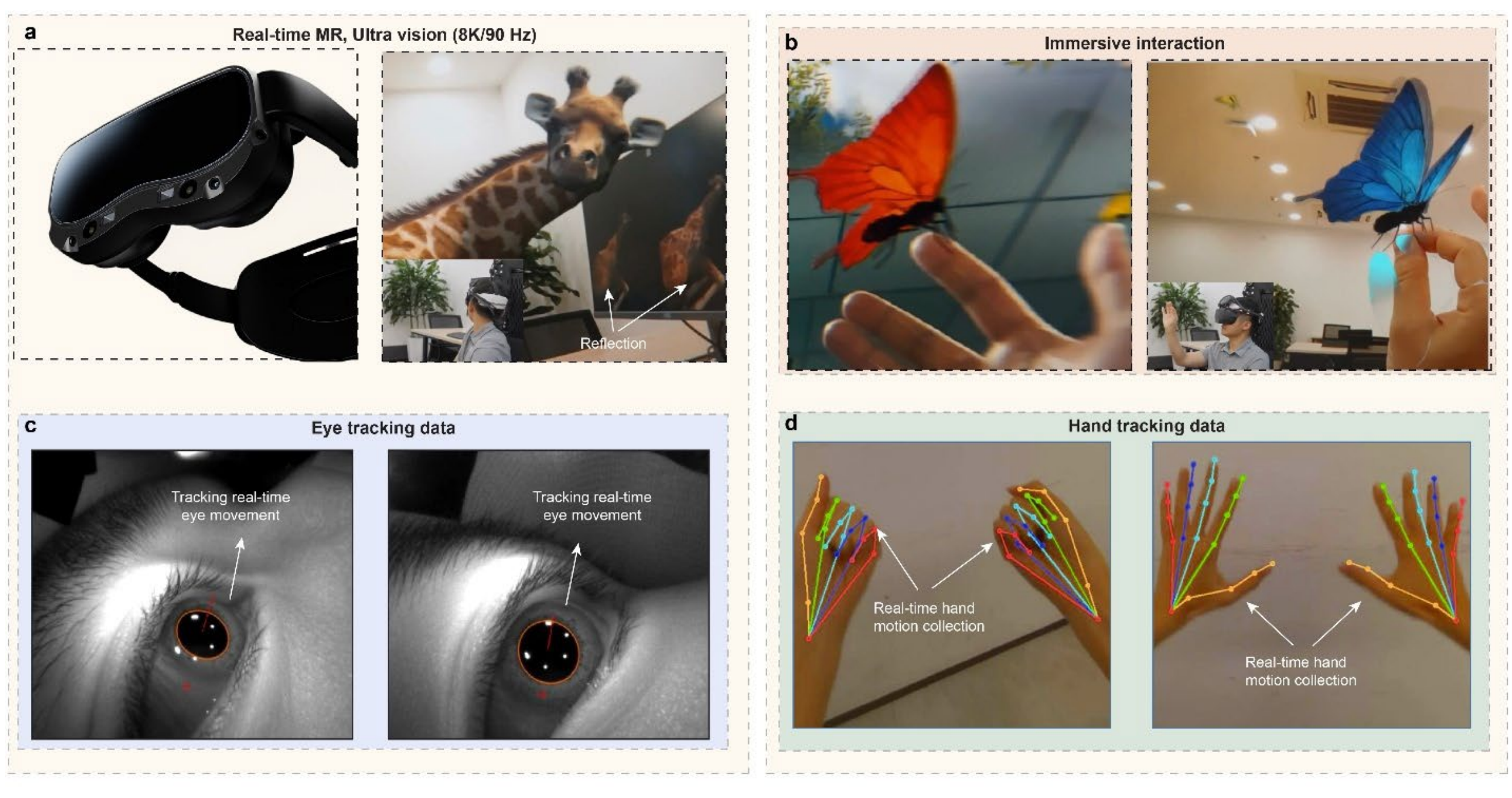

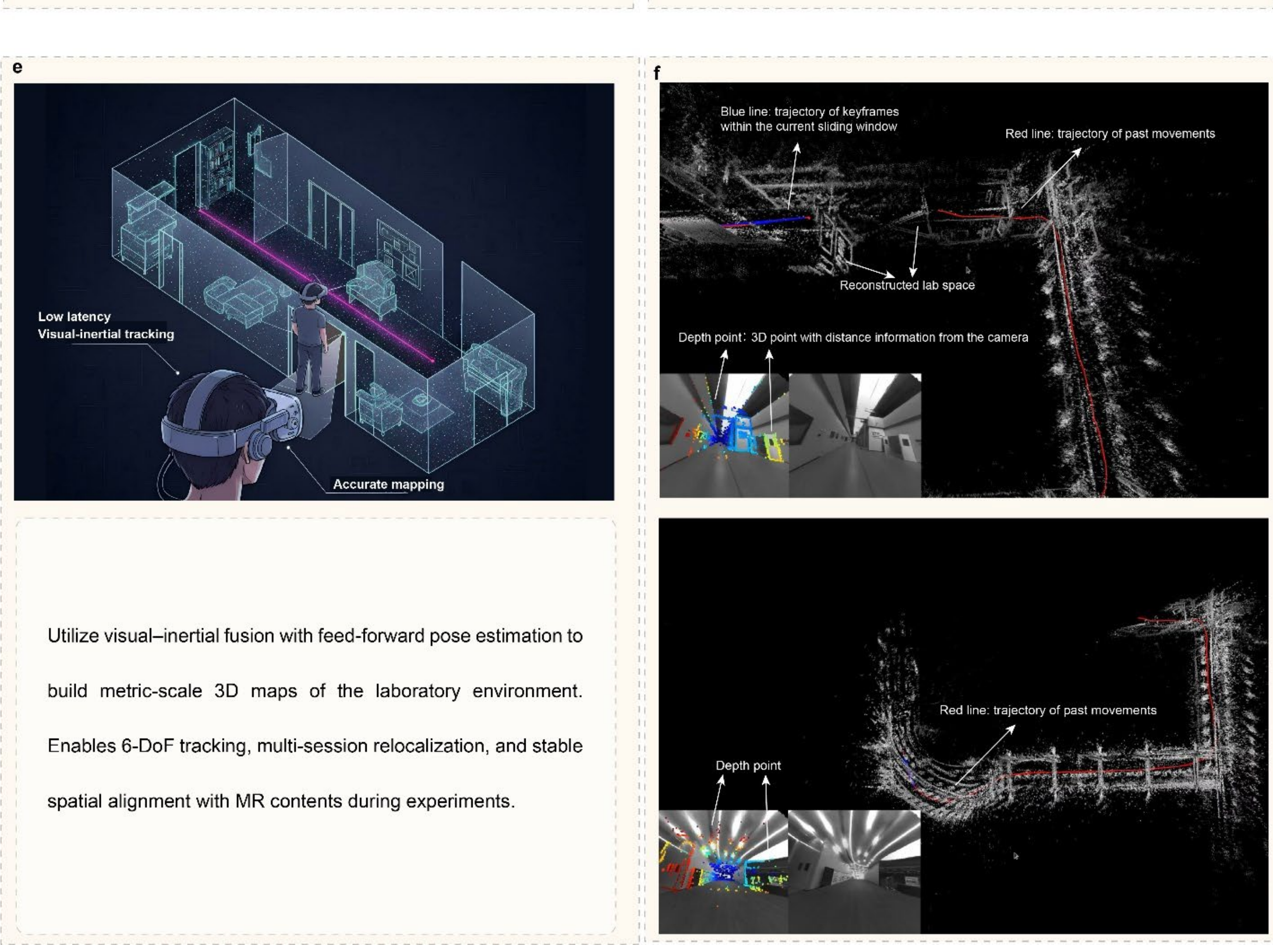

**Extended Data Fig. 2 | Multimodal perception and spatial mapping for MR interaction and laboratory reconstruction. a,** Real-time MR system integrating 8K/90 Hz video see-through and augmented reality visualization that enables seamless understanding of experimental scenes and bidirectional feedback between the human researchers and APEX system. **b,** Representative

immersive interactions from the human researcher's perspective. **c,** Eye-tracking subsystem providing real-time measurement of gaze direction and ocular motion for visual attention analysis within the MR workspace. **d,** Hand-tracking subsystem capturing 3D hand poses and finger trajectories to support natural, gesture-based interaction within the MR workspace. **e,** Visual-inertial simultaneous localization and mapping (VI-SLAM) framework for real-time 3D reconstruction of fabrication and manufacturing environments. **f,** Metric-scale reconstruction results showing keyframe trajectories (blue) and historical trajectories (red), with the visualization of depth maps derived from multi-view camera data. This system supports six degrees-of-freedom (6-DoF) tracking, multi-session re-localization, and robust spatial alignment, enabling stable MR-based robust spatiotemporal reconstruction for agentic experiment guidance.

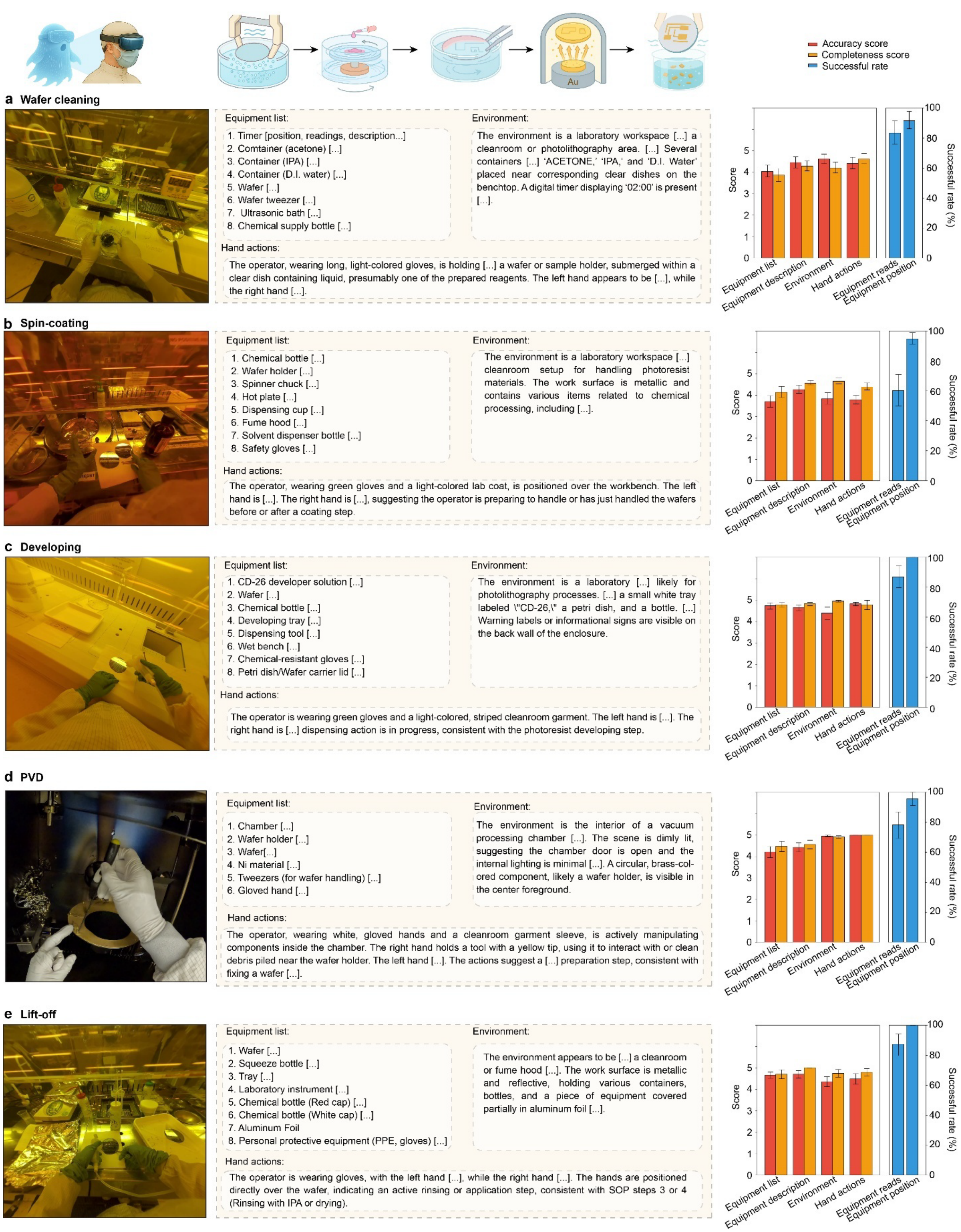

**Extended Data Fig. 3 | Scene understanding and procedural annotation across representative microfabrication steps. a–e,** Bar plots show the accuracy of APEX system's scene understanding capability. The APEX system interprets multimodal inputs from diverse fabrication processes,

including wafer cleaning (**a**), spin-coating (**b**), developing (**c**), PVD (**d**), and lift-off (**e**). For each process, MR frames are decomposed into three structured descriptions: an equipment list with equipment identity, live readings, and spatial position; environment summary describing the workspace state; and research hand actions. Parameter readings and positions in APEX system-generated descriptions are assigned a score of 1 or 0 if they are correct or incorrect, respectively, and the accuracy and completeness of the equipment list, environment summary and hand actions descriptions are evaluated by a fabrication expert on a scale from 1–5 (**Supplementary Note 1**) in a fully blinded setup. In **a–e,** bar plots indicate mean ± s.e.m. ($n \geqslant 23$ MR experiment frames). Together, these structured annotations provide a quantitative evaluation of APEX system's scene understanding capability and its supply reproducible data for downstream procedural reasoning and continual model retraining in scientific experimentation and intelligent manufacturing.

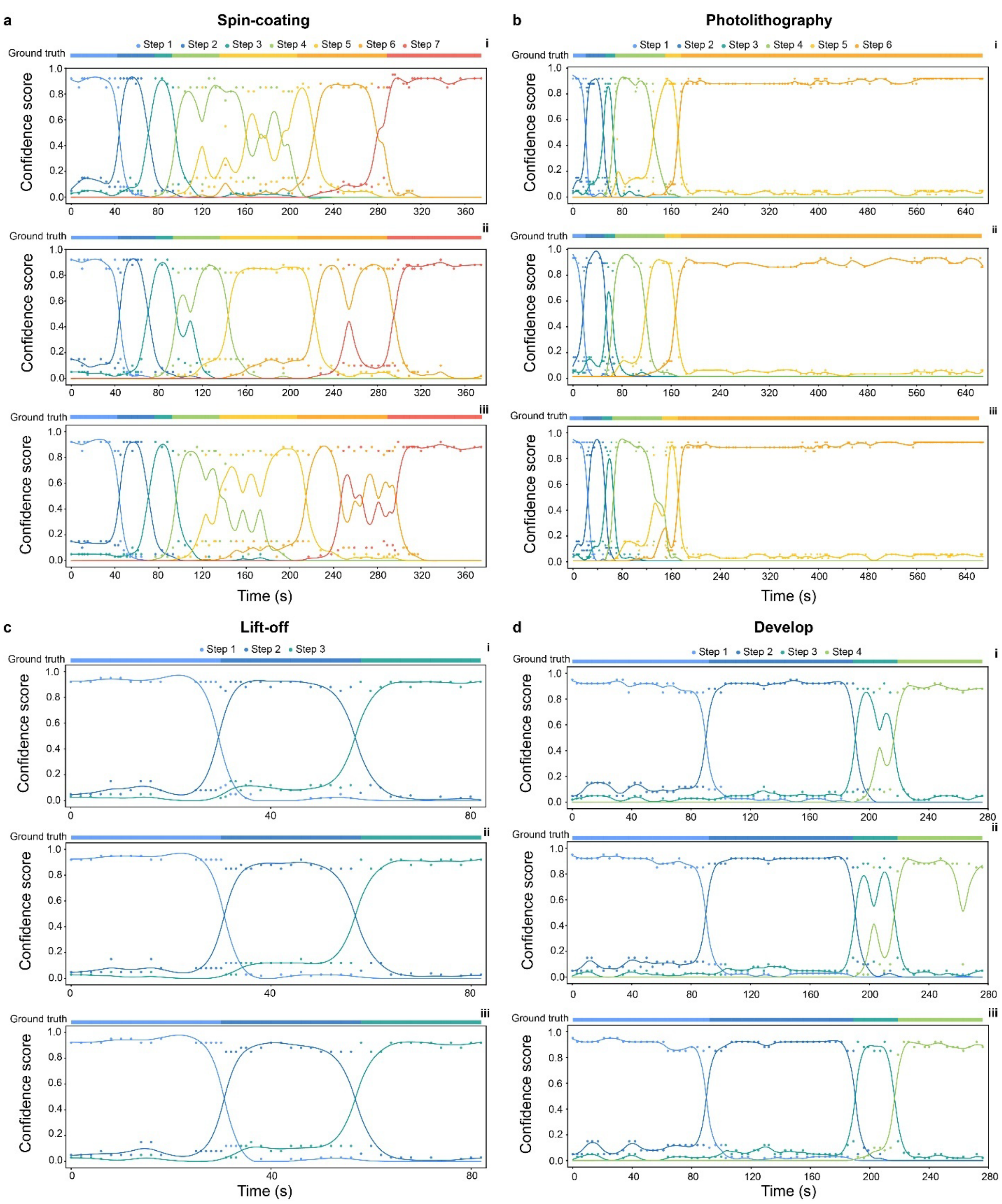

**Extended Data Fig. 4 | Representative temporal traces of step-tracking self-reported confidence from different microfabrication processes. a,** Spin-coating; **b,** Photolithography; **c,** Lift-off**; d,** Develop. APEX system sustains high self-reported confidence across successive steps, reflecting robust alignment between perceived human actions and the planned SOP across different microfabrication processes. Each dot represents one MR experiment frame; temporal prediction is

visualized by traces smoothed over the dots using radial basis functional interpolation with a thin-plate spline kernel ($\lambda$ = 100).

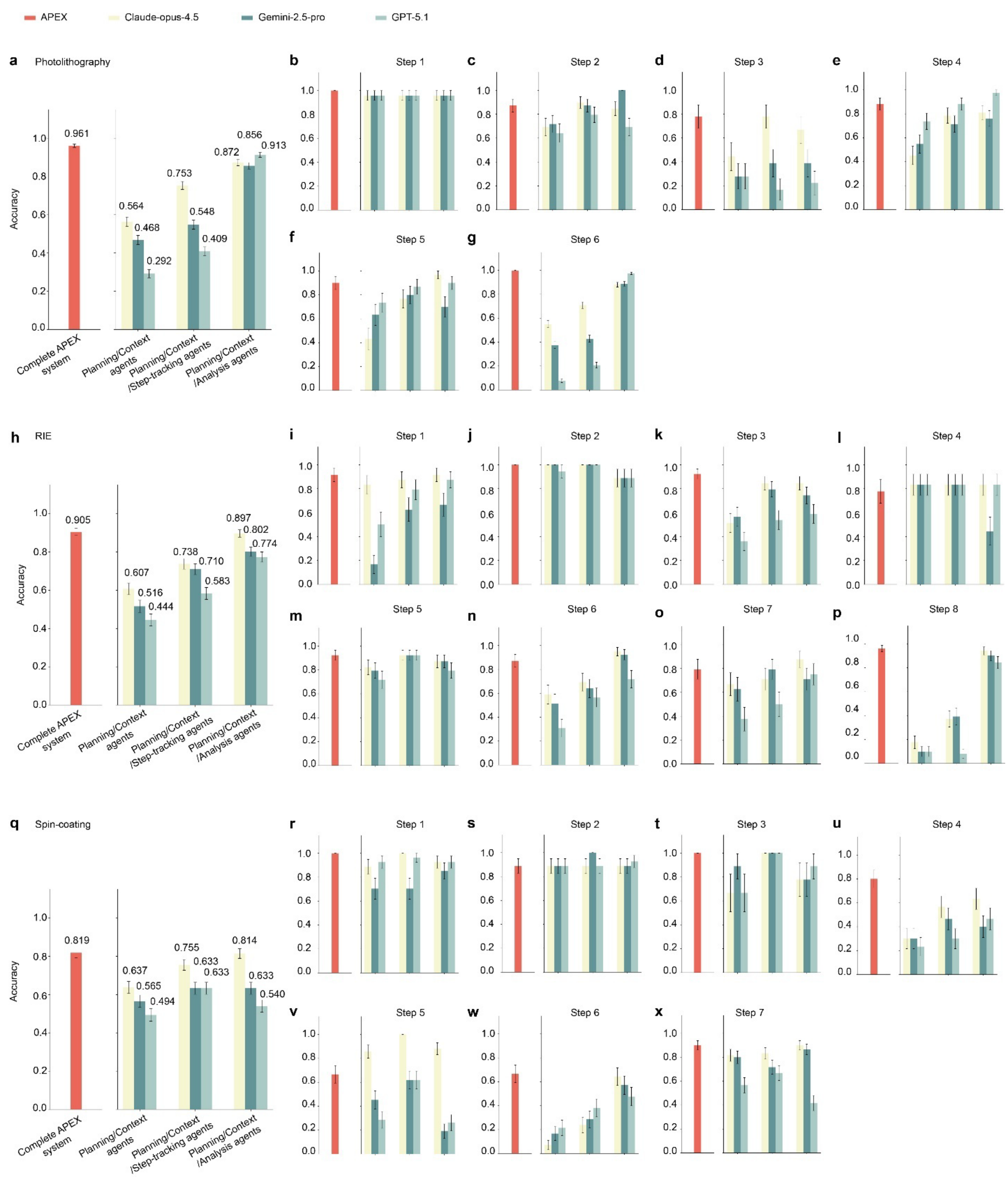

**Extended Data Fig. 5 | Step-tracking benchmark across different microfabrication processes.** Bar plots quantify the step-tracking accuracy in **a–g,** Photolithography; **h–p;** RIE; and **q–x,** spin-coating. The complete APEX system (Planning/Context/Step-tracking/Analysis agents) performance was compared against its ablated configurations: Planning/Context agents, Planning/Context/Step-tracking agents, and Planning/Context/Analysis agents. Panels (**a**), (**h**), and (**q**) show the mean step-tracking accuracy across all procedural steps for microfabrication

processes. Panels (**b**)–(**g**), (**i**)–(**p**), and (**r**)–(**x**) show stepwise tracking accuracy for procedural steps in photolithography, RIE, and spin-coating, respectively. In **a–x**, bar plots indicate mean ± s.e.m. ($n \geqslant 9$ MR experiment frames).

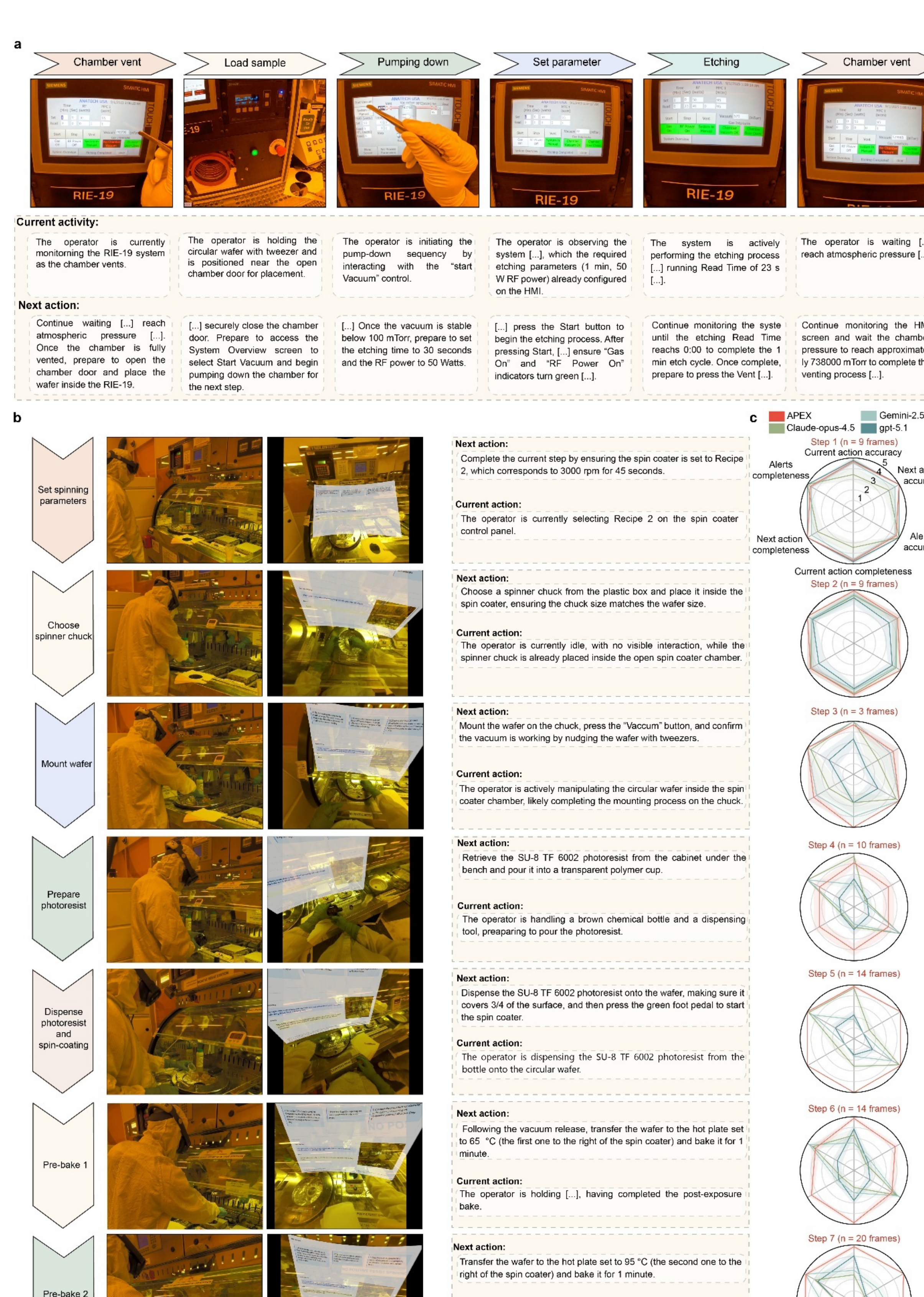

**Extended Data Fig. 6 | Seamless human–AI collaboration for scientific experiments and intelligent manufacturing in microfabrication. a,** Complete RIE process performed by a first-time user under MR guidance. Stepwise overlays and predictive instructions allow consistent procedural accuracy comparable to experts, highlighting APEX system's potential as a scalable physical AI platform for reproducible, human–AI co-fabrication. **b,** Real-time collaboration during the spin-coating process. APEX system continuously perceives the operator's actions and surrounding context through the MR input, providing adaptive, stepwise feedback for parameter setup, photoresist dispensing, coating, and baking. At each step, the system interprets the current action, predicts the next appropriate step, and issues contextual alerts when procedural deviations are detected, all without interrupting the operator's workflow. **c,** Quantified completeness and accuracy of human–AI collaboration generated by APEX and other LLMs in a representative spin-coating process. Alerts predicted next actions, and current-action descriptions generated by APEX and baseline LLMs were evaluated by a fabrication expert on a scale from 1–5 (**Supplementary Note 1**) in a fully blinded setup. Solid lines in the radar plot indicate the mean; shaded error band represents the s.e.m. ($n \geqslant 3$ MR experiment frames).

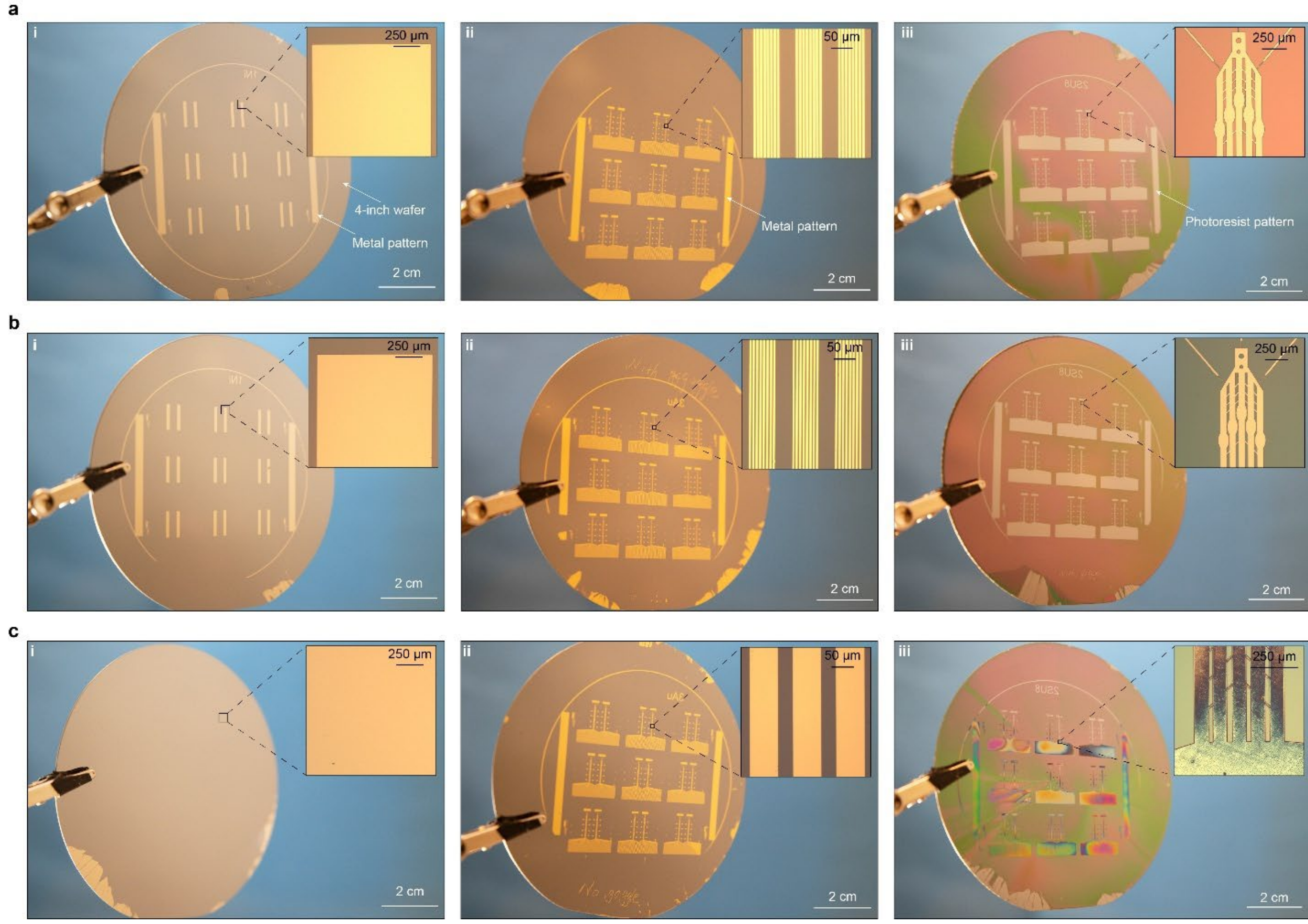

**Extended Data Fig. 7 | Representative photolithographic patterns developed by experts and novices under different guidance conditions. a,** Patterns developed by an expert, serving as a reference for optimal photolithographic fabrication. **b,** Patterns developed by a novice assisted by the APEX system. **c,** Patterns developed by a novice without expert and APEX system guidance. Wafer-scale images and corresponding magnified optical images of representative regions are shown for each condition.

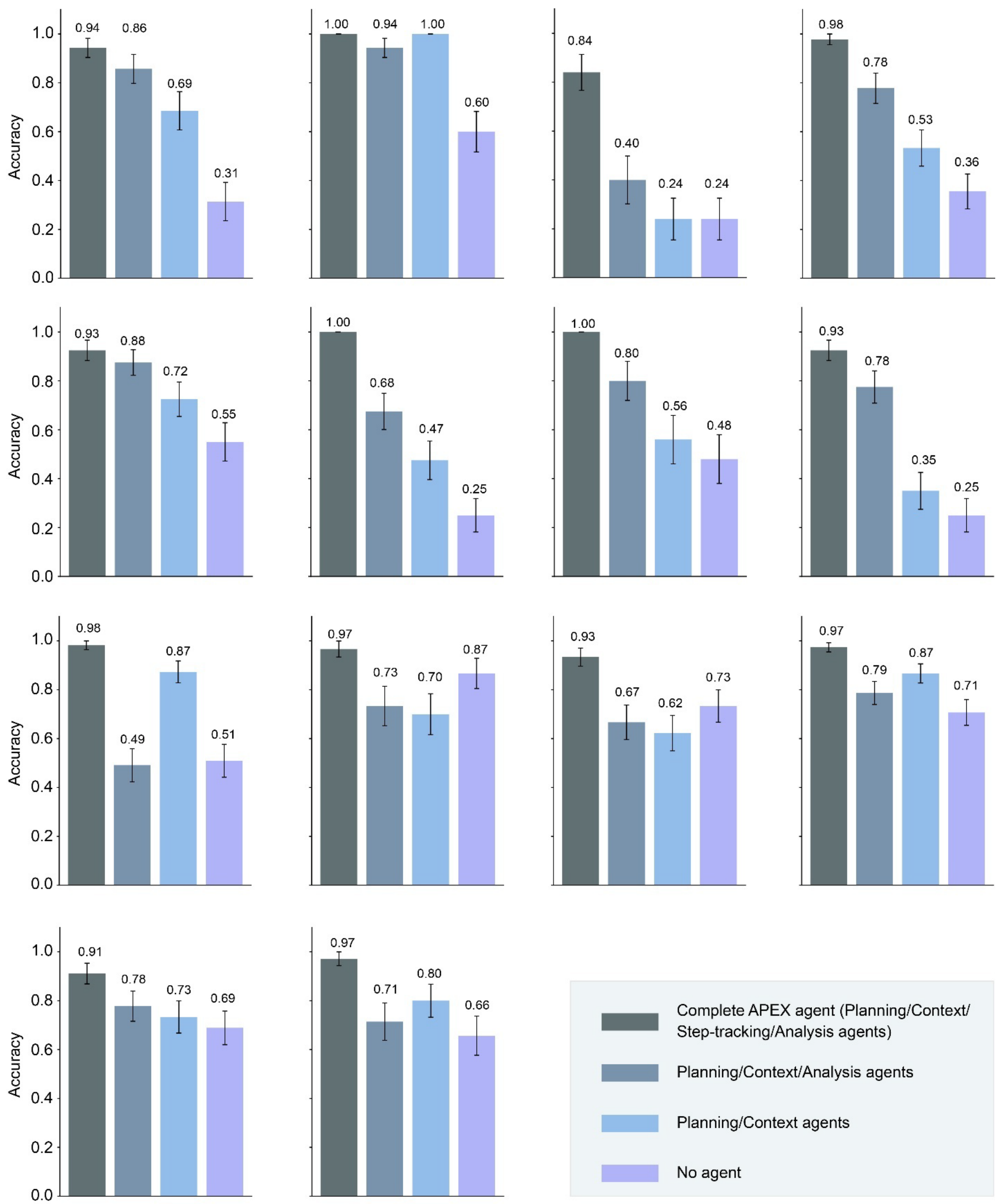

**Extended Data Fig. 8 | Error detection accuracy of APEX system under 14 representative RIE scenarios.** Error detection accuracy across 14 representative RIE scenarios under different agent configurations, including the complete APEX system, Planning/Context/Analysis agents, Planning/Context agents, and one that is fully ablated. Bar plots present the mean ± s.e.m. ($n \geq 25$ MR experiment frames).

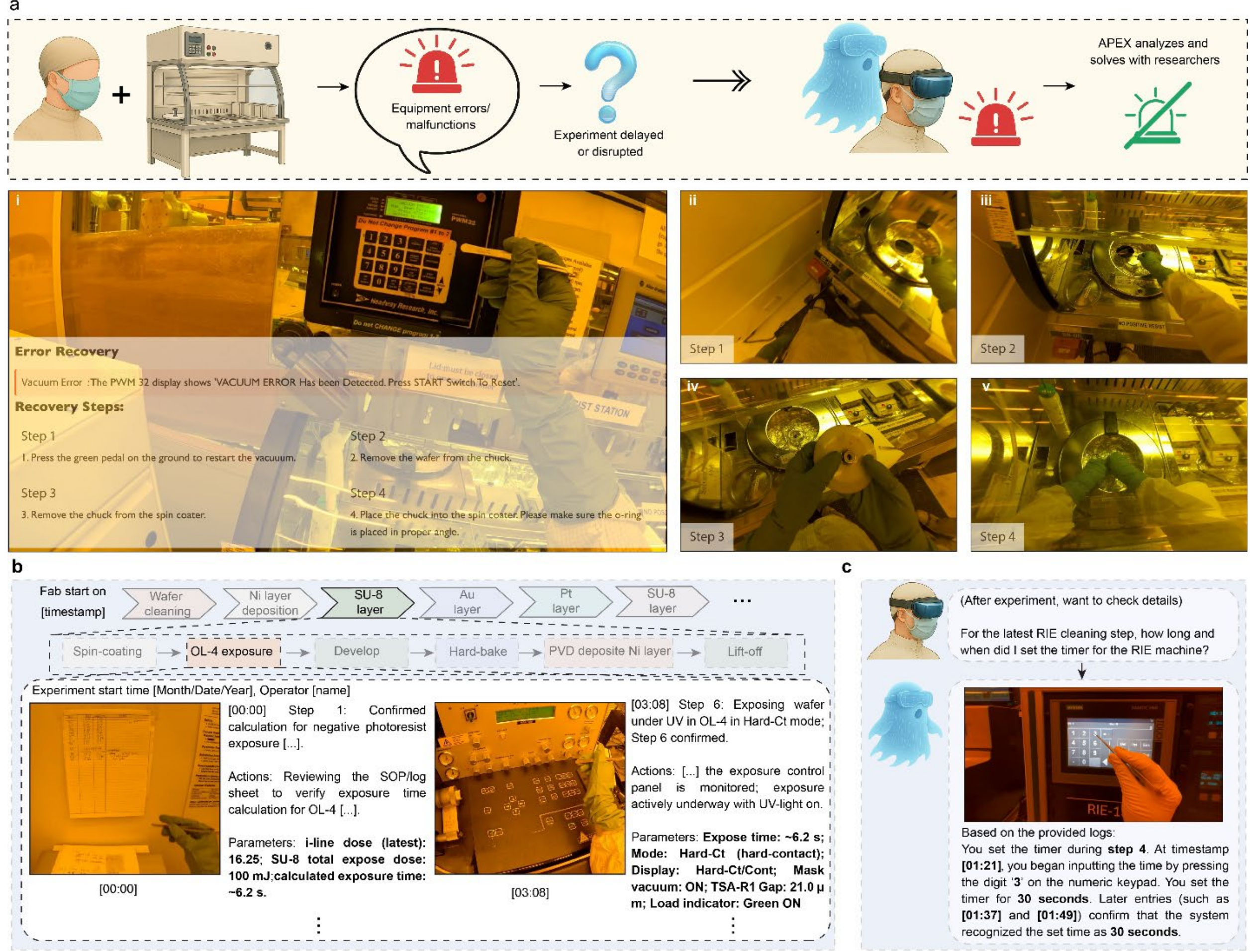

**Extended Data Fig. 9 | APEX system autonomously recovers equipment error and enhances experiment traceability. a,** Top, schematic illustration of a representative equipment error scenario in which an unexpected malfunction disrupts an ongoing microfabrication experiment, leading to uncertainty and procedural interruption. APEX system identifies the error context and collaborates with researcher to diagnose and resolve the issue. Bottom, representative example of vacuum error recovery during a spin-coating process. The APEX system detects the vacuum error, represents stepwise recovery instructions through the MR interface (i), and guides the user through successive recovery actions (ii–v), including equipment reset, wafer removal, chuck repositioning, and process resumption. This example illustrates how APEX system supports real-time error handling and recovery during microfabrication workflows. **b,** Autonomous experiment logging. APEX system automatically records parameters, timestamps, environmental snapshots, and actions into a structured digital log, forming a comprehensive, searchable record for reproducibility and quality assurance. **c,** Human–AI questions and answers reflection. APEX system supports post-experiment review, allowing users to query key process steps, parameter deviations, and outcomes with the generated structured experimental memory using natural language.

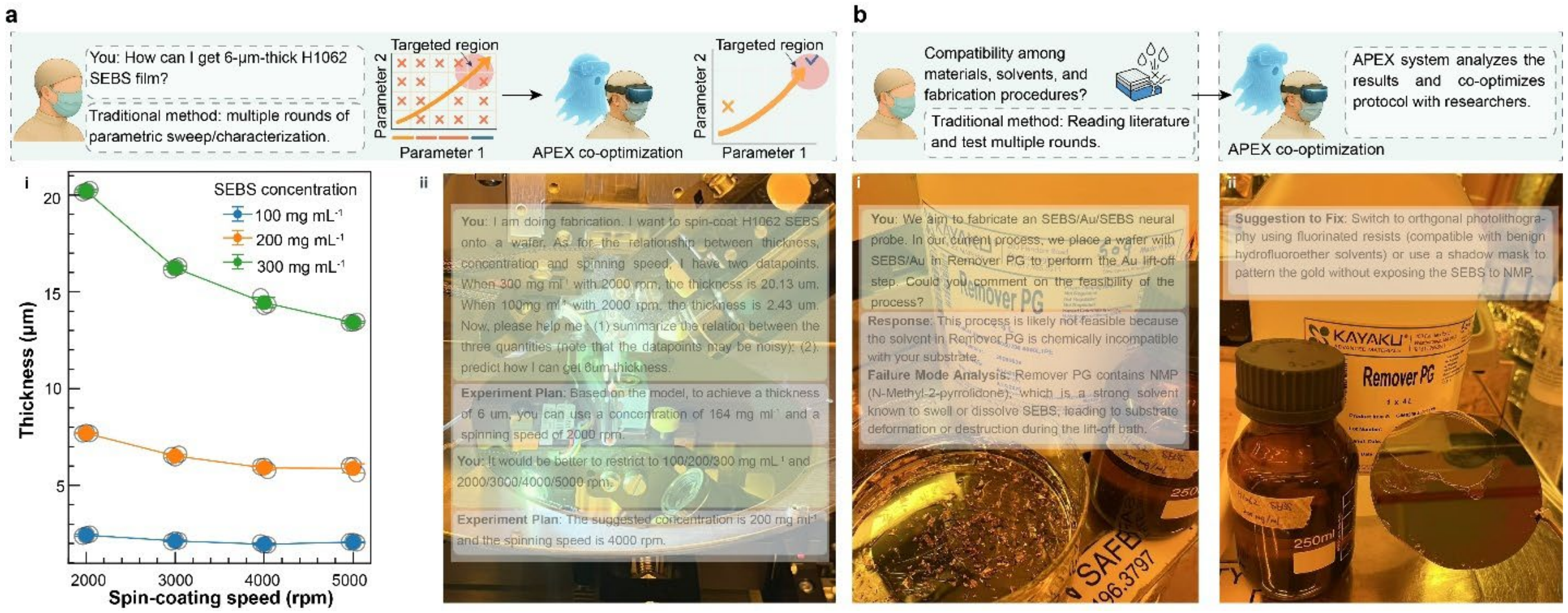

**Extended Data Fig. 10 | APEX system assists in the development and feasibility assessment of new protocols in collaboration with users. a**, APEX-assisted acceleration of parametric optimization, enabling efficient exploration of fabrication parameters through guided experimental design and analysis. The traditional method relies on repeated manual experimentation to empirically derive the relationship between spin-coating speed and SEBS film thickness (i), while the APEX-assisted user can directly obtain the necessary parameter (ii). **b**, APEX-assisted evaluation and resolution of material, solution, and process compatibility issues, supporting and accelerating protocol optimization during device development (i,ii). In **a,** colored line-and-dot plots represent mean; colored error bars represent the s.e.m.; each concentric dot represents an individual film thickness at different locations on the same wafer; $n$ = 3 measurements per spin-coating speed and concentration.